\documentclass[twoside]{article}

\usepackage[accepted]{aistats2025}
\usepackage[round]{natbib}

\usepackage{placeins}
\usepackage{microtype}
\usepackage{graphicx}
\usepackage{subfigure}
\usepackage{booktabs} 
\usepackage{epigraph}
\usepackage{wrapfig}
\newcommand{\enum}[1]{\label{eq:#1}}
\newcommand{\eref}[1]{(\ref {eq:#1})}

\newcommand{\be}{\begin{equation}}
\newcommand{\ee}{\end{equation}}

\usepackage{hyperref}
\usepackage{array}

\usepackage{amsmath}
\usepackage{amssymb}
\usepackage{mathtools}
\usepackage{amsthm}

\usepackage[capitalize,noabbrev]{cleveref}

\theoremstyle{plain}
\newtheorem{theorem}{Theorem}[section]
\newtheorem{proposition}[theorem]{Proposition}

\theoremstyle{definition}

\theoremstyle{remark}

\usepackage[utf8]{inputenc} 
\usepackage[T1]{fontenc}    
\usepackage{hyperref}       
\usepackage{url}            
\usepackage{booktabs}       
\usepackage{amsfonts}       
\usepackage{nicefrac}       
\usepackage{microtype}      
\usepackage{xcolor}         

\begin{document}

%

%

\twocolumn[

\aistatstitle{GD doesn't make the cut: Three ways that non-differentiability affects neural network training}

\aistatsauthor{Siddharth Krishna Kumar}
\aistatsaddress{siddharthkumar@upwork.com}

]

\begin{abstract}
This paper critically examines the fundamental distinctions between gradient methods applied to non-differentiable functions (NGDMs) and classical gradient descents (GDs) for differentiable functions, revealing significant gaps in current deep learning optimization theory. We demonstrate that NGDMs exhibit markedly different convergence properties compared to GDs, strongly challenging the applicability of extensive neural network convergence literature based on $L-smoothness$ to non-smooth neural networks. Our analysis reveals paradoxical behavior of NDGM solutions for $L_{1}$-regularized problems, where increasing regularization counterintuitively leads to larger $L_{1}$ norms of optimal solutions. This finding calls into question widely adopted $L_{1}$ penalization techniques for network pruning. We further challenge the common assumption that optimization algorithms like RMSProp behave similarly in differentiable and non-differentiable contexts. Expanding on the Edge of Stability phenomenon, we demonstrate its occurrence in a broader class of functions, including Lipschitz continuous convex differentiable functions. This finding raises important questions about its relevance and interpretation in non-convex, non-differentiable neural networks, particularly those using ReLU activations. Our work identifies critical misunderstandings of NDGMs in influential literature, stemming from an overreliance on strong smoothness assumptions. These findings necessitate a reevaluation of optimization dynamics in deep learning, emphasizing the crucial need for more nuanced theoretical foundations in analyzing these complex systems.
\end{abstract}

\section{Introduction}

Gradient Descent (GD) and its variants \citep{duchi2011adaptive,kingma2014adam,lydia2019adagrad,mcmahan2010adaptive,shi2021rmsprop,tieleman2012lecture,zhang2018improved} have been instrumental in advancing image and language processing over the last two decades. These algorithms have gained widespread adoption in neural network training due to their ease of implementation, high scalability, and ability to iteratively approach stationary points where the loss function's gradient vanishes \citep{bertsekas1997nonlinear}.

However, an important consideration has emerged in the field: despite being originally designed for differentiable loss functions, GDs are routinely applied to minimize loss functions of non-differentiable neural networks. These non-differentiable loss functions possess a unique characteristic—they are differentiable almost everywhere. Consequently, when gradient descents are applied to these non-differentiable neural networks (which we term non-differentiable gradient methods or NGDMs), they rarely encounter non-differentiable points during training. This property, coupled with recent findings suggesting NGDMs converge to stationary points under mild regularity conditions \citep{bolte2021conservative,davis2020stochastic}, has led to a widespread assumption in the field: that non-differentiability in neural networks is inconsequential (see Section 6.1.5 of \citet{stevens2020deep}, for instance). Notably, three of the most widely cited surveys on optimization methods for neural networks \citep{bottou2018optimization, le2011opti, ruder2016overview} do not discuss non-differentiability, despite the ReLU—arguably the most ubiquitous activation function in neural networks—being non-differentiable.

This assumption has not only rekindled interest in gradient descent dynamics within complex loss landscapes \citep{ahn2022understanding,dauphin2014identifying,reddi2019convergence}, but has also led to a trend where theoretical papers develop results for continuously differentiable functions and then claim universality by running simulations on non-differentiable neural networks (Experiment 6 in \citet{ahn2022understanding}, Section 3.1 in \citet{ma2022multiscale}, Section 6 in \citet{zhang2022adam}).

Our paper examines this fundamental assumption, demonstrating that it leads to significant inaccuracies in several highly referenced papers and texts regarding neural network training dynamics. We analyze three critical aspects of the training process, making the following key contributions:

\begin{itemize}
\item \textbf{Convergence Analysis of ReLU Networks:} We show that NGDMs converge more slowly than GDs, suggesting that many convergence rate claims made under the assumption of $L-smoothness$ have limited applicability to NGDM properties.

\item \textbf{Solutions to the LASSO problem:} We demonstrate that NGDMs do not produce sparse solutions for the LASSO problem, even in the simplest case of the $L_1$ penalized linear model. This finding contrasts with prevalent understanding. Moreover, we show that NGDMs can yield unexpected results, such as producing solutions with larger $L_1$ norms when using larger LASSO penalties. Importantly, we prove that variants like RMSProp and NDGM with momentum behave differently from vanilla NDGM, challenging the common assumption that these algorithms converge to similar solutions in non-differentiable settings.

\item \textbf{The Edge of Stability:} We demonstrate that the edge of stability conjecture does not hold, even for gradient descents on all convex differentiable neural networks. Furthermore, we provide a counter-example to the claim made in \citet{ma2022multiscale} that subquadratic behavior around the local minimum is responsible for the Edge of Stability phenomenon.
\end{itemize}

In a broader context, our analysis underscores the importance of the assumptions we make in understanding these complex systems. For NGDMs in particular, our results suggest a need to reconsider how we approach these methods, challenging the common assumption of their similarity to GDs and calling for a more nuanced theoretical foundation in deep learning optimization.

\section{Preliminaries}

In the following three sections, we demonstrate how disregarding the non-differentiability of the loss function significantly limits our understanding of various aspects of training dynamics. Each section is largely self-contained, including its own literature review, problem setup, and analysis. Throughout this paper, our experiments with deep networks consistently employ convolutional neural network (CNN) architectures similar to VGG16 \citep{simonyan2014very}, chosen for its widespread use in the papers we reference. Our experiments have relatively low computational demands, with the most expensive run completing within 2 hours on a single GPU. This section introduces background and terminology relevant to all subsequent sections.

\paragraph{NDGM vs. GD:} The core of our analyses lies in the structural differences between GDs and NDGMs. For a continuously differentiable loss function, $f(\beta)$, the well-studied GD sequence \citep{bertsekas1997nonlinear,boyd2004convex} is described by the recursion
\be 
\beta_{t+1} = \beta_{t} - \alpha\nabla f(\beta_{t}), \enum{prel1}
\ee 
where $\alpha$ is the learning rate, and $\beta_{0}$ is a randomly initialized starting point. 

For a continuous non-differentiable loss function, $g(\beta)$, the NDGM recursion is given by
\be 
\Tilde{\beta}_{t+1} = \Tilde{\beta}_{t} - \alpha\widetilde{\nabla} g(\Tilde{\beta}_{t}), \enum{prel2}
\ee 
where $\widetilde{\nabla} g(\Tilde{\beta}_{t}) = \nabla g(\Tilde{\beta}_{t})$ if the gradient exists at $\Tilde{\beta}_{t}$, else it is a heuristic measure \footnote{\url{https://rb.gy/g74av}}. The deep learning literature widely assumes that the dynamics described by the recursions in \eref{prel1} and \eref{prel2} are nearly identical.

\paragraph{Convex non-differentiable loss functions:} We demonstrate the issues with this assumption using various convex loss functions. Most of our analysis focuses on the penalized single-layer neural network regression problem with a non-positive response vector, which \citet{kumar2023analytical} has shown to be convex with the all-zeros vector as its unique global minimizer. The loss function for this problem is:

\begin{align}
    L_{1}(&\beta;{\bf{Z}},{\bf{y}},\lambda_{1},\lambda_{2}) = \nonumber \\
    &\|{\bf{y}} - \max(0,{\bf{Z}}\beta)\|^{2} + \lambda_{1}\|\beta\|_{1} + \lambda_{2}\|\beta\|_{2}^{2},
    \label{eq:prel0}
\end{align}

where $\beta$ is a $P \times 1$ parameter vector, ${\bf{Z}}$ is an $N \times P$ data matrix with $i^{th}$ column $z_{i}$, ${\bf{y}}$ is a $N \times 1$ non-positive response vector, $\lambda_{1} \geq 0$ is the LASSO penalty, $\lambda_{2} \geq 0$ is the ridge penalty, and $\lambda_{1} + \lambda_{2} > 0$. We denote the learning rate by $\alpha$, and the $i^{th}$ entry in the parameter vector after $k$ iterations as $\beta_{k}[i]$, with $0 \leq i \leq P-1$.  

\paragraph{The subgradient method:} Our analysis relies heavily on the observation that for convex non-differentiable loss functions, the NDGM recursion in \eref{prel2} effectively implements the subgradient method \citep{bertsekas1997nonlinear,shor2012minimization}. The theoretical properties of this method are well-studied in optimization literature, with key findings summarized in the lecture notes of \citet{boyd2003subgradient} and \citet{tibshirani2015subgradient}. The subgradient method and GD recursions have different properties, which play a crucial role in our discussions. We will reference relevant portions of these lecture notes as they become pertinent to our analysis.

\paragraph{Constant, diminishing, and reducing step sizes:} Our formulations in \eref{prel1} and \eref{prel2} use a constant learning rate. We explain this choice by recalling definitions from optimization literature (see Chapter 1 of \citet{bertsekas1997nonlinear}). The training regimen in \eref{prel1} is called a constant step-size regime, where GD is initialized randomly and iterated with a constant learning rate until the loss stops decreasing. Constant step-size GDs approach the stationary point but rarely converge to it unless the loss function has desirable properties like $L-smoothness$ (Proposition 1.2.3 in \citet{bertsekas1997nonlinear}). For provable convergence in more general settings, studies often use the diminishing step-size regime, characterized by:
\be
\beta_{t+1} = \beta_{t} - \alpha_{t}\nabla f(\beta_{t}), \enum{prel3}
\ee
where $\lim_{t\rightarrow\infty} \alpha_{t} = 0$, and $\sum_{t} \alpha_{t} \rightarrow \infty$. Convergence results in this regime provide theoretical guarantees under ideal conditions. In practice, diminishing step-size GDs use stopping criteria to end training. We term the combination of diminishing step-size and stopping criteria commonly used in neural network literature the "reducing" step-size regime.

In this regime, training proceeds as follows: starting with weight $\beta_{0}$, GD runs for $N_{0}$ iterations with learning rate $\alpha_{0}$, followed by $N_{1}$ iterations with reduced rate $\alpha_{1} < \alpha_{0}$, and so on, until satisfactory results are achieved. This involves chaining together $T$ constant step-size GD runs, with the $(i+1)^{th}$ run initialized at $\beta_{u}$ with learning rate $\alpha_{i}$, where $u = \sum_{j=0}^{i-1} N_{j}$. While theoretical guarantees of the diminishing step-size regime may not be realized, the limiting dynamics in this regime are those of a constant step-size GD, with learning rates and weight initializations chosen from the $T^{th}$ run. Therefore, we exclusively study the constant step-size problem, assuming appropriate weight initializations and learning rates.

\section{NDGM and convergence analysis} \label{sec:ndgm-convergence}

Recent theoretical work by \citet{davis2020stochastic} has shown that in the diminishing step-size regime, NDGMs converge to a local optimal solution in the theoretical limit under fairly general conditions. However, these conditions are rarely satisfied in practice. For instance, Assumption A.3 in \citet{davis2020stochastic} requires that the sum of the learning rates should diverge in the diminishing step size regime. While this is a standard assumption for GD analysis, it's often violated in practice. The prevalent step-size reduction in most training scripts involves decaying the learning rate by a constant multiplicative factor at predefined intervals (e.g., StepLR in PyTorch\footnote{\url{https://rb.gy/pub4s}}), resulting in a convergent geometric series of learning rates, directly contradicting this assumption.

For differentiable functions, we know that even with a convergent geometric series of learning rates, the GD sequence approaches the vicinity of the optimal solution. However, it's unclear if this holds true for NDGM sequences. More precisely, can we assert that the dynamics of NDGM and GD sequences are identical? 
\paragraph{NDGM and GD dynamics are not identical.} 
We demonstrate that the dynamics of NDGM differ qualitatively from those of gradient descent (GD) by arriving at a contradiction. Suppose that the sequence $\{\beta_k\}_{k=0}^\infty$ generated by running NDGM on the objective $L_2(\beta; {\bf Z}_1, {\bf q}_1) = L_1(\beta; {\bf Z}_1, {\bf q}_1, 0.001, 0)$ exhibits the same behavior as a GD sequence. If, after some index $m$, the sequence produces a monotone non-increasing loss (which we ensure using an adaptive step size—the learning rate is reduced by a factor of 0.9 every time the loss increases; see Appendix \ref{app:convanalysis} for the full code), then the tail of the sequence $\{\beta_{m+i}\}_{i=0}^{\infty}$ satisfies the conditions of the Capture Theorem. 

\begin{figure}[!htbp]
    \centering
    \includegraphics[width=0.95\linewidth]{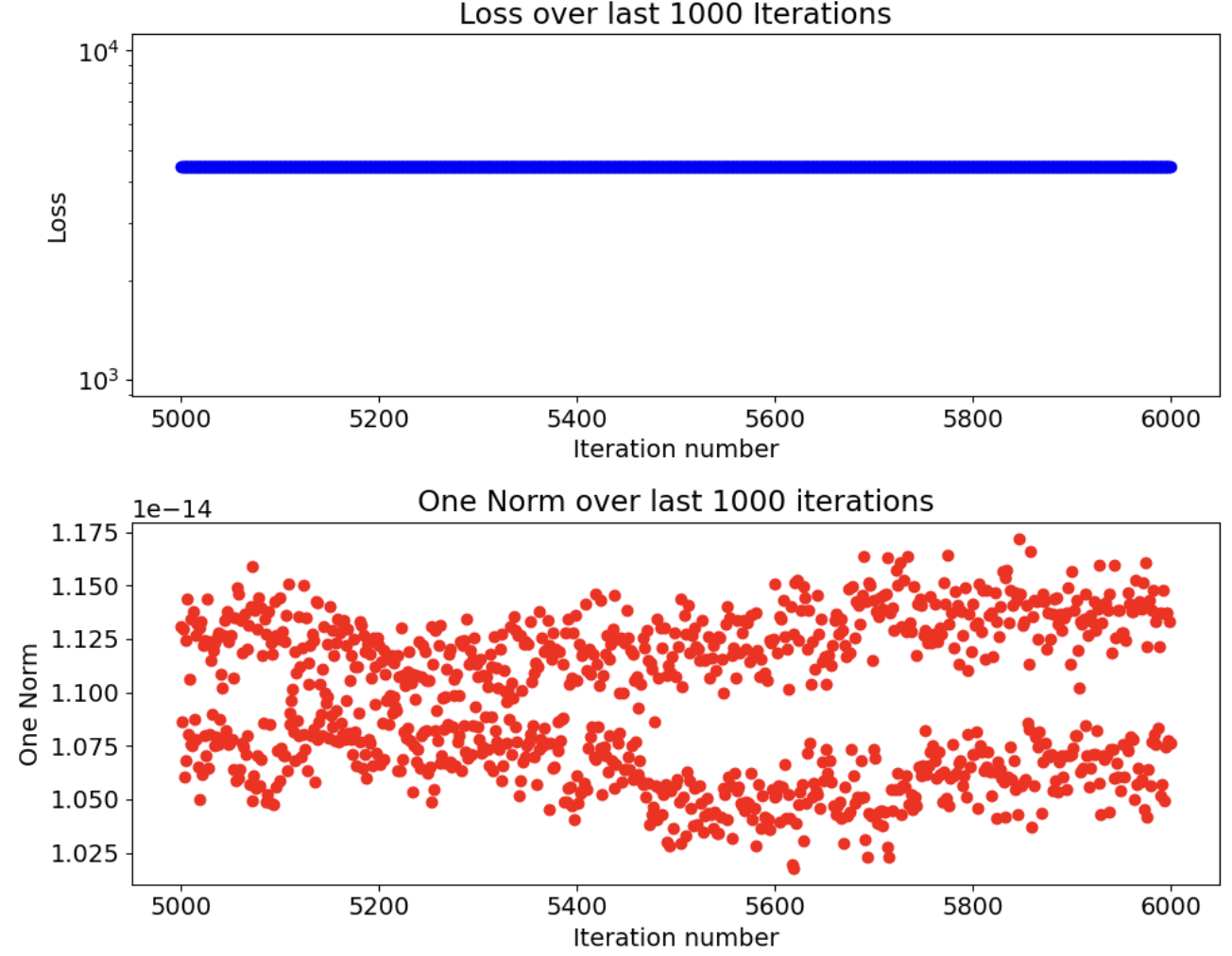}
    \caption{Although loss is monotone non-increasing (upper panel), the one norm of the weights does not capture (lower panel).}
    \label{fig:relu_plot_one_norm}
\end{figure}

Informally, the theorem states that when GD is applied to a differentiable function and produces a monotone non-increasing sequence of losses in a neighborhood of an isolated local minimum $\beta^*$, the iterates must eventually be ``captured'': they remain within that neighborhood and get progressively closer to $\beta^*$ in norm. Formally, there exists an $N_0$ such that for all $q \geq p \geq N_0$, $||\beta_{q} - \beta^{*}|| \leq ||\beta_{p} - \beta^{*}||$. In our case, where $\beta^* = 0$, we enforce eventual monotonicity of the loss using the adaptive step-size rule. However, as shown in Figure \ref{fig:relu_plot_one_norm}, while the loss eventually becomes monotonic, the norm of the weight vector does not—yielding a contradiction. This failure of capture is expected: as noted in \citet{boyd2003subgradient}, ``Unlike the ordinary gradient method, the subgradient method is not a descent method; the function value can (and often does) increase.'' Even when we explicitly enforce monotonic loss, NDGM's dynamics remain fundamentally different from those of GD.

\paragraph{Convergence analysis and L-smoothness:}

Given the distinct dynamics of GDs and NDGMs, we must reconsider the assumptions appropriate for convergence analysis in neural network training. Many analyses in the literature \citep{bottou2018optimization,zhang2022adam,defossez2020simple,you2019large} assume that the loss function $f(\beta)$ is continuously differentiable and $L-smooth$, i.e., there exists a constant $L$ such that:

\be 
||\nabla f(x) - \nabla f(y)|| \leq L||x - y|| \enum{lsmooth}
\ee 

for all $x$, $y$ in the domain of $f$. However, this assumption fails for non-differentiable functions where the gradient doesn't exist at various points. One might consider salvaging this assumption by limiting its validity to a local context around the local optimal solution.

Unfortunately, even this approach is untenable. As \citet{laurent2018multilinear} state, "local minima of ReLU networks are generically nondifferentiable. They cannot be waved away as a technicality, so any study of the loss surface of such network must invoke nonsmooth analysis". This observation aligns with established ideas in convex optimization. For instance, the development and application of second-order cone programs (SOCPs) for convex non-smooth problems was motivated by similar considerations \citep{boyd2004convex}\footnote{See 26:30 to 29:00 in \url{https://rb.gy/ab0aeh}}. In both SOCPs and ReLU networks, local minima often occur at points of non-differentiability, necessitating non-smooth analysis.

Moreover, assuming $L-smoothness$ significantly overestimates the convergence speed of NDGMs. For an error level of $\epsilon$, GD applied to an $L-smooth$ convex loss function requires $O(1/\epsilon)$ steps for convergence. In contrast, NDGM applied to a convex non-differentiable loss function satisfying the Lipschitz condition:

\be 
||g(x) - g(y)|| \leq L||x - y|| \enum{lips}
\ee 

takes $O(1/\epsilon^{2})$ steps (see Section 7.1.3 of \citet{tibshirani2015subgradient} for the result and a visual illustration of how much slower $O(1/\epsilon^{2})$ is than $O(1/\epsilon)$). Thus, the extensive literature on GD convergence analysis for $L-smooth$ functions likely has limited relevance to the convergence dynamics of modern non-smooth neural networks. Generally, we expect smooth neural networks to converge faster than non-smooth ones.

\paragraph{Experiments with NDGM convergence:}

We conduct two sets of experiments on CNNs of varying depths to study how activation smoothness affects convergence (see Appendix~\ref{app:relugelucomp} for code). For each architecture, we compare the non-smooth ReLU with the smooth and asymptotically equivalent GELU activation \citep{hendrycks2016gaussian}. We also include Softplus activations with $\beta=100$ and $\beta=1000$ to illustrate how varying curvature affects convergence, although these are not asymptotically equivalent to ReLU and are not directly comparable.

\begin{figure}[!htbp]
\centering
\begin{minipage}{0.48\textwidth}
  \centering
  \includegraphics[width=0.48\textwidth]{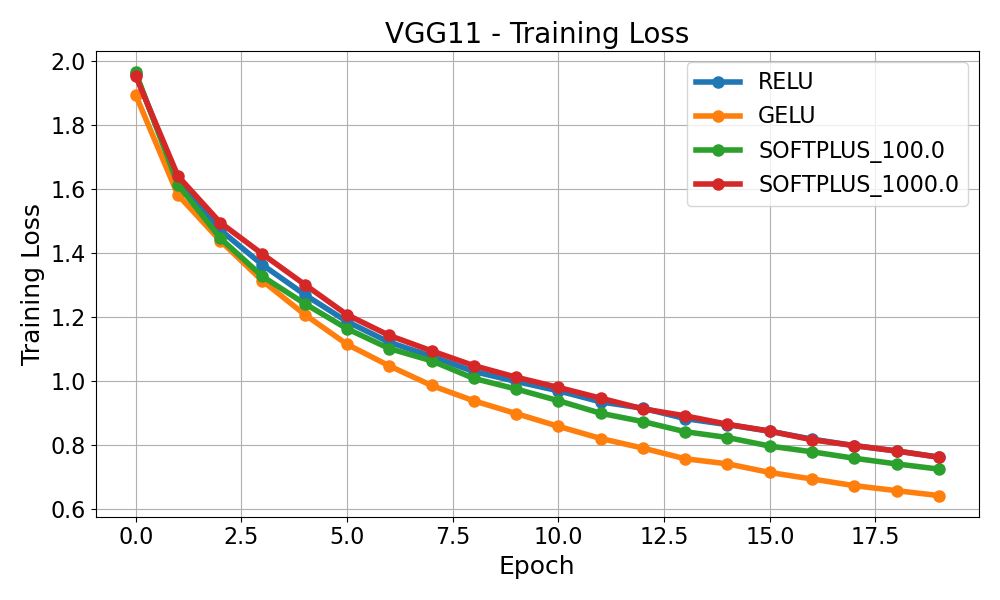}\hfill
  \includegraphics[width=0.44\textwidth]{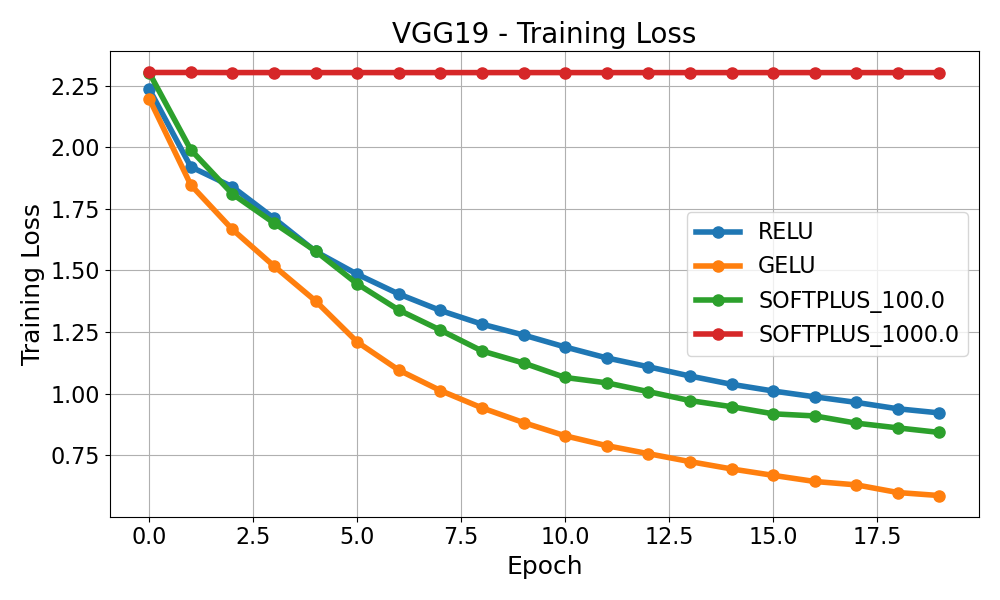}
  \vspace{-0.3cm}
  \includegraphics[width=0.44\textwidth]{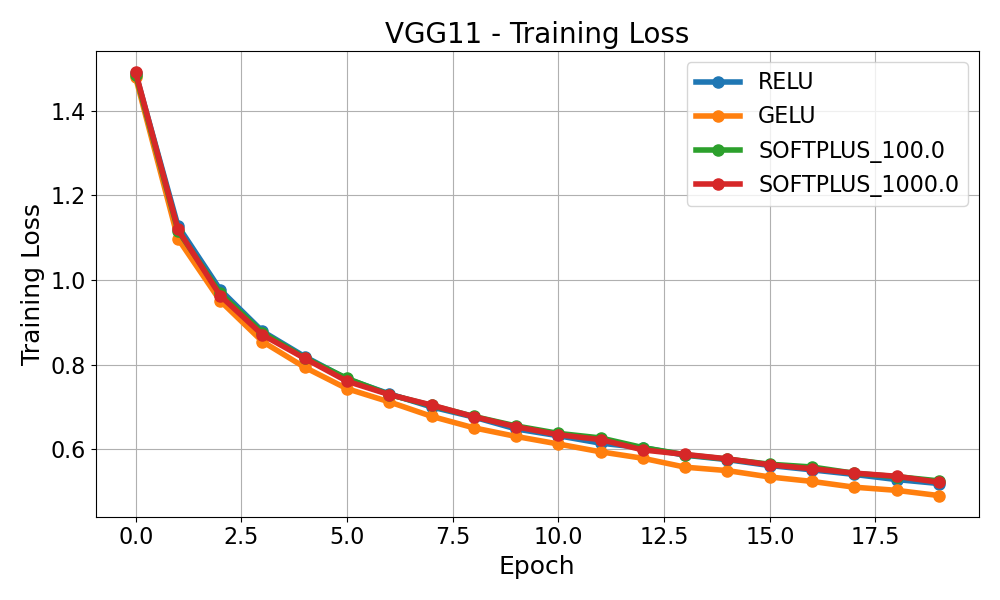}\hfill
  \includegraphics[width=0.44\textwidth]{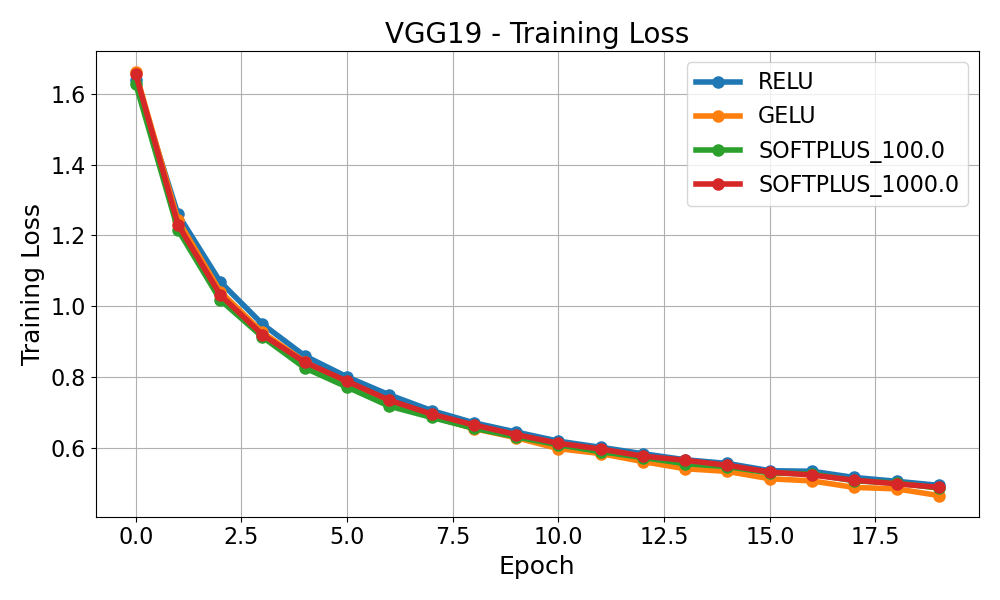}
\end{minipage}
\caption{Training loss for CNNs with different activation functions: ReLU (blue), GELU (orange), Softplus-$\beta$=100 (green), and Softplus-$\beta$=1000 (red). Top: without batch normalization. Bottom: with batch normalization. Left column: VGG11. Right column: VGG19.}
\label{fig:convergence}
\end{figure}

Figure~\ref{fig:convergence} shows that without batch normalization, smoother activations converge faster—particularly as depth increases. This aligns with the findings of \citet{hendrycks2016gaussian}, who reported faster convergence with GELU across diverse architectures. With batch normalization, these differences disappear, highlighting its strong regularizing effect.

For the Softplus variants, we observe that $\beta=1000$ converges more slowly than $\beta=100$, consistent with the intuition that sharper activations make optimization harder. However, we also find that $\beta=1000$ is slower than ReLU, while $\beta=100$ is faster—despite both being smooth. We do not have a principled explanation for this behavior, and in the absence of a well-defined equivalence relation between Softplus and ReLU, we interpret these results as exploratory.

\paragraph{L-smoothness and the geometry of ReLU networks:} The widely used $L-smoothness$ assumption, which ensures that the gradient of the loss function does not change too rapidly with respect to the parameter vector (Section 4.1 in \citet{bottou2018optimization}), does not hold for ReLU networks. To illustrate this, consider the 1-D case with $q(x) = \text{max}(0,x)$. In this case, $\lim_{h \rightarrow 0^{+}} (1/2h)(\nabla q(h) - \nabla q(-h)) = \infty$, indicating that the gradient changes arbitrarily quickly at the origin. This necessitates the use of non-smooth analysis for studying these losses. The issue, as previously discussed, is that non-differentiable functions have kinks corresponding to non-differentiable points, and local minima frequently lie at these kinks. Therefore, gradient-based approaches developed for differentiable functions are insufficient since they cannot access these points of interest (see Appendix \ref{app:cvxanal} for geometric intuitions into this fact). The inability of gradient descent methods to effectively maneuver non-smooth functions and the associated pathologies form a recurring theme in this paper. $L-smoothness$ is one of the strongest assumptions after convexity, guaranteeing the existence and boundedness of the first and second derivatives—criteria unmet by most loss functions in neural network literature. To emphasize this point, we show in a later section that results derived for smooth networks may not generalize even to all Lipschitz continuous convex differentiable networks.

\paragraph{Limitations of our work:} A common limitation in the current and subsequent sections is that we do not analyze the dynamics of the neural network training process as it is often conducted in practice. This is necessitated by the fact that neural networks often involve dropout layers, which leave the loss function discontinuous, let alone non-differentiable, and are thus beyond the scope of our study. We do not reiterate this limitation in every subsequent section.

\paragraph{Extending our results to the general neural network setting:} Our analysis in \ref{eq:prel0} focuses on the specific case of non-positive response vectors, which allows for a convex formulation. Extending these results to general single-layer neural networks with arbitrary response vectors, let alone multi-layer networks, is challenging due to the resulting non-convex and non-differentiable loss functions. Currently, there is no known way to rigorously analyze such networks in their full generality. This limitation is not unique to our work—it is a fundamental challenge in the field. Our paper's key contribution is to rigorously demonstrate that this challenge cannot be addressed by simply extrapolating results from the analysis of smooth functions to non-smooth settings, as is often done in the literature. We specifically caution against the common practice of deriving results for $L-smooth$ networks and then claiming generality based on simulations on architectures like VGG16 or ResNet. This approach, while prevalent, fails to capture the fundamental differences between smooth and non-smooth optimization that we highlight. Our findings are crucial for the field, as they expose the limitations of current methods and underscore the need for more rigorous and appropriate techniques in analyzing ReLU networks and other non-smooth architectures.

\paragraph{The way forward:} In this section, we show that NDGMs are slower to converge than GDs. Fortunately, most recent explorations into image and language models primarily leverage the transformer architecture \citep{vaswani2017attention} with the differentiable GELU activation function \citep{hendrycks2016gaussian}. As a consequence, future iterations of neural networks may predominantly employ pure GDs. Therefore, it is crucial for the research community to establish a consensus on a set of assumptions that accurately represent these complex systems before meaningful progress can be achieved.

\section{NDGM and the LASSO penalty}\label{sec:ndgmlasso}

Zero weights are crucial for model compression \citep{blalock2020state, han2015deep, han2015learning, li2016pruning}, and inducing sparsity by penalizing the $L_1$ norm of model weights is a common approach. While the statistics literature employs specialized algorithms for problems involving penalized $L_1$ norms \citep{efron2004least, friedman2008sparse, mazumder2012graphical, tibshirani1996regression, tibshirani2005sparsity, zou2005regularization}, neural networks often use the penalized $L_1$ norm with arbitrary non-convex functions, assuming that NDGM results in a near-sparse optimal solution \citep{bengio2012practical,goodfellow2016regularization,scardapane2017group}. However, this assumption is not satisfied in practice, as evidenced by user struggles in PyTorch forums\footnote{\url{https://rb.gy/kh0qr}}\footnote{\url{https://rb.gy/2vsry}}\footnote{\url{https://rb.gy/6dfli}}. Given the LASSO's critical applications \citep{ghosh2021efficient, li2011bayesian, li2012overview, ogutu2012genomic}, understanding the reliability of the NDGM solution for the LASSO problem is essential.

\paragraph{NDGM solutions for the LASSO problem are not reliable:} We demonstrate that NDGM produces unreliable solutions even for the simple case of the LASSO penalized linear model. Specifically, we show that increasing the LASSO penalty can result in a NDGM solution with a larger $L_{1}$ norm, completely defeating the purpose of the penalty.
We first analytically demonstrate the unreliability for the simplest LASSO problem, 
\be 
L_{3}(\beta) = L_{1}(\beta;0,0,\lambda_{1},0) = \lambda_{1}||\beta||_{1} , \enum{las01}
\ee 
with $\lambda_{1} > 0$, and then demonstrate the same issues for the general LASSO problem through simulations. We begin by showing that for any non-zero learning rate, $\alpha$, the NDGM sequence for \eref{las01} will be non-sparse with probability 1.

\begin{proposition}
    \label{prop:lassooverall}
    Let $\{\beta_{t}\}_{t=0}^\infty$ denote the sequence generated by running NDGM on \eref{las01}, with a constant learning rate, $\alpha$, and the entries in $\beta_{0}$ being uniformly sampled from $[-1,1]$. Then $\beta_{k}$ will have all non-zero entries with probability 1 for all $k > 0$. Furthermore, there exists an integer $N_{0}$, such that for every $N \geq N_{0}$, we can write

    \be
    \beta_{N + m}[k] = \begin{cases}
                        \gamma_{k} - \alpha\lambda_{1} & \text{ if m is odd} \\ 
                       \gamma_{k} & \text{ if m is even},
                  \end{cases} \enum{lassooverall}
\ee
where $\gamma_{k} \in (0,\alpha\lambda_{1})$ for all $0\leq k\leq P-1$.
\end{proposition}
\begin{proof}
    See Appendix \ref{app:lassooverall} for details
\end{proof}

Equation \eref{lassooverall} suggests that with the same learning rate and initialization, a larger $\lambda_{1}$ can lead to a NDGM solution with a larger $L_{1}$ norm, defeating the purpose of the LASSO penalty. To illustrate this, we note that in the 2D case, with weights initialized as $\beta_{0} = [0.5053, 0.5053]$, $\alpha = 0.01$, and $\lambda_{1} = 1$, the NDGM sequence oscillates between $[0.0053, 0.0053]$ and $[-0.0047, -0.0047]$. However, with $\lambda_{1} = 100$, and the same values of $\beta_{0}$ and $\alpha$, the solution oscillates between $[0.5053, 0.5053]$ and $[-0.4947, -0.4947]$. Therefore at convergence, the $L_{1}$ norm is at most 0.0106 for $\lambda_{1} = 1$, but at least 0.9894 for $\lambda_{1} = 100$. Furthermore, reducing the step-size to $\alpha = 0.001$ after apparent convergence does not resolve this paradox. With $\lambda_{1} = 1$, the sequence oscillates between $[0.0003, 0.0003]$ and $[-0.0007, -0.0007]$, while with $\lambda_{1} = 100$, it oscillates between $[0.0053, 0.0053]$ and $[-0.0947, -0.0947]$. Appendix \ref{app:lassogeneral} shows via simulation that this paradox holds for NDGM solutions of the standard $L_{1}$ regularized linear problem.

\paragraph{The subgradient method and step-size choices:} The counter-intuitive behavior seen in the example above can be explained using the subgradient method. Since the loss function described in equation \eref{las01} is Lipschitz continuous (equation \eref{lips}) with $L = \lambda_{1}$, we have from \citet{boyd2003subgradient} (Section 2 on the Constant Step Size) that 
\be
    \lim_{k \rightarrow \infty} ||\beta_{k}||_{1} < \alpha\lambda_{1}. \enum{lasserr}
\ee 
Therefore, for the same value of $\alpha$, the larger value of $\lambda_{1}$ will result in a larger error in estimation, which is precisely what we are seeing above. This illustration also highlights the contrast between the diminishing and reducing step size regimes outlined in the preliminaries. Under the diminishing step-size regime, $\alpha \rightarrow 0$ in the theoretical limit, and therefore, Proposition \ref{prop:lassooverall} guarantees that the optimal solution for \eref{las01} will be 0 for all $\lambda_{1} > 0$. On the other hand, in the reducing step-size regime, $\alpha > 0$ at the end of training, and therefore, the error in estimation becomes a critical determinant of the optimal value.

\paragraph{Connections to other NDGM variants:} Conventional wisdom in deep learning theory suggests that different NDGM variants converge at varying rates but eventually reach similar near-optimal values (Figure 8.5 of \citet{goodfellow2016deep}, slides 64 onwards in \url{https://rb.gy/ojzpgm} \footnote{These slides are from one of the most widely used courses for self teaching; see 1:01:00 - 1:03:00 of \url{https://rb.gy/kaj976} where this precise point is made}). However, we analytically show that these claims are incorrect even for the toy LASSO problem described in \eref{las01} (see Appendices \ref{app:ndgmvariants} and \ref{app:dltheory} for details). Furthermore, our analysis reveals two interesting facts which cannot be explained by the analysis of differentiable functions: 1) the RMSProp sequence for \eref{las01} is agnostic of the value of $\lambda_{1}$ used in the analysis, and 2) there is a phase transition between the RMSProp and NDGM training curves based on the magnitude of $\lambda_{1}$. These findings highlight the importance of incorporating the non-differentiability of regularizers like the $L_1$ norm when analyzing the behavior of NDGM variants.

\paragraph{Experiments with deep networks:}
The seminal work on network pruning by \citet{han2015learning} proposes using $L_1$ regularization to encourage weights to approach 0, facilitating the pruning process. This approach has been widely adopted in the literature \citep{wen2016learning,xiao2019autoprune,yang2019structured} and industry\footnote{See the recommendation of the regularizer in the Training config section of \url{https://rb.gy/y7o7hw}}. To examine the general applicability of this method, we conducted two sets of experiments finetuning a pre-trained VGG16 model on CIFAR-10 using $L_1$ penalized cross-entropy loss with a constant learning rate of $\alpha = 0.003$ for 200 epochs. We used $\lambda_{1} = 0.0002$ in the first set and $\lambda_{1} = 2$ in the second, running three parallel trials within each set using RMSProp, SGD, and Adam. 

\begin{figure}[!htbp]
\centering
\begin{minipage}{0.48\textwidth}
  \centering
  \includegraphics[width=0.48\textwidth]{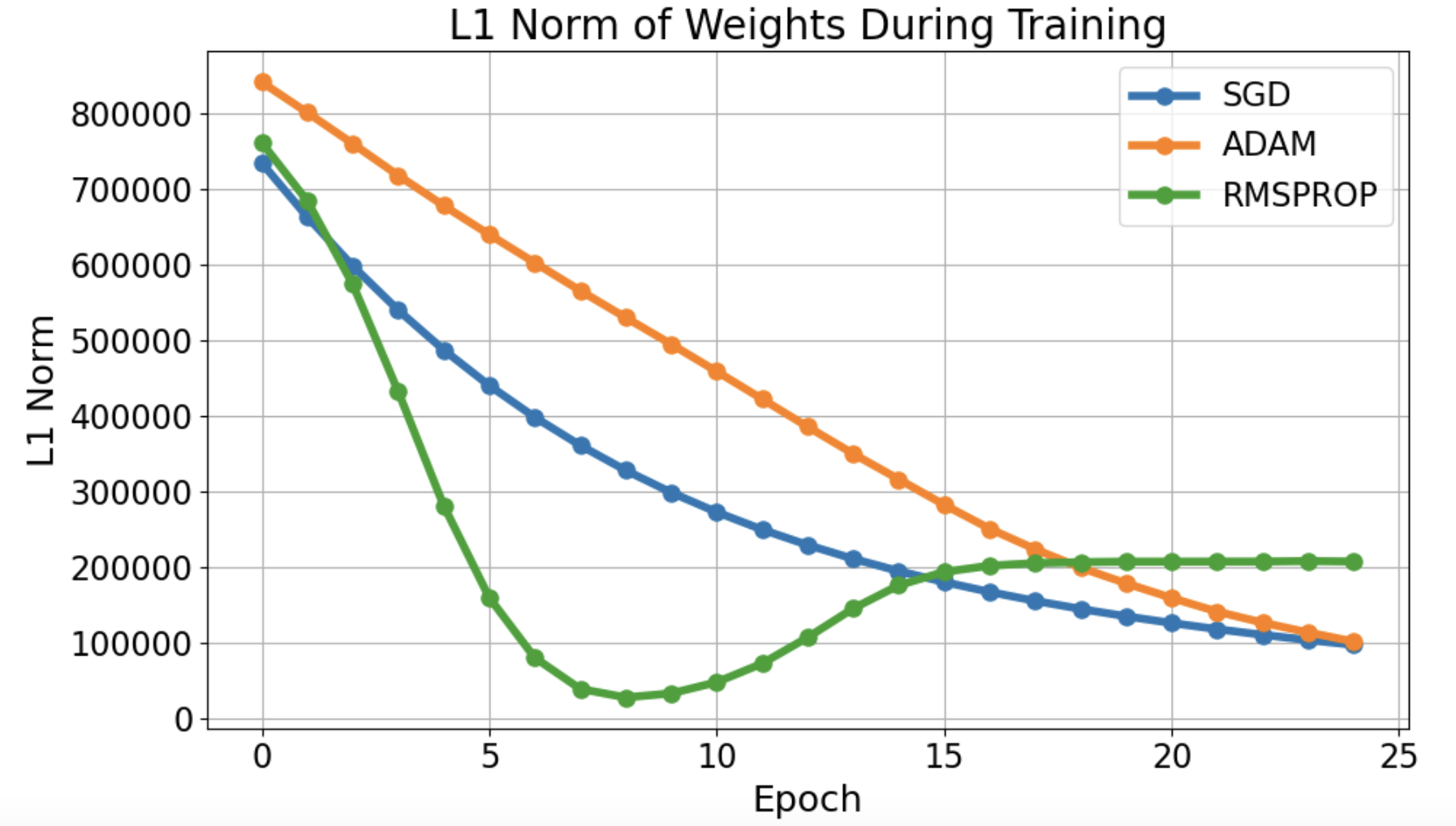}\hfill
  \includegraphics[width=0.48\textwidth]{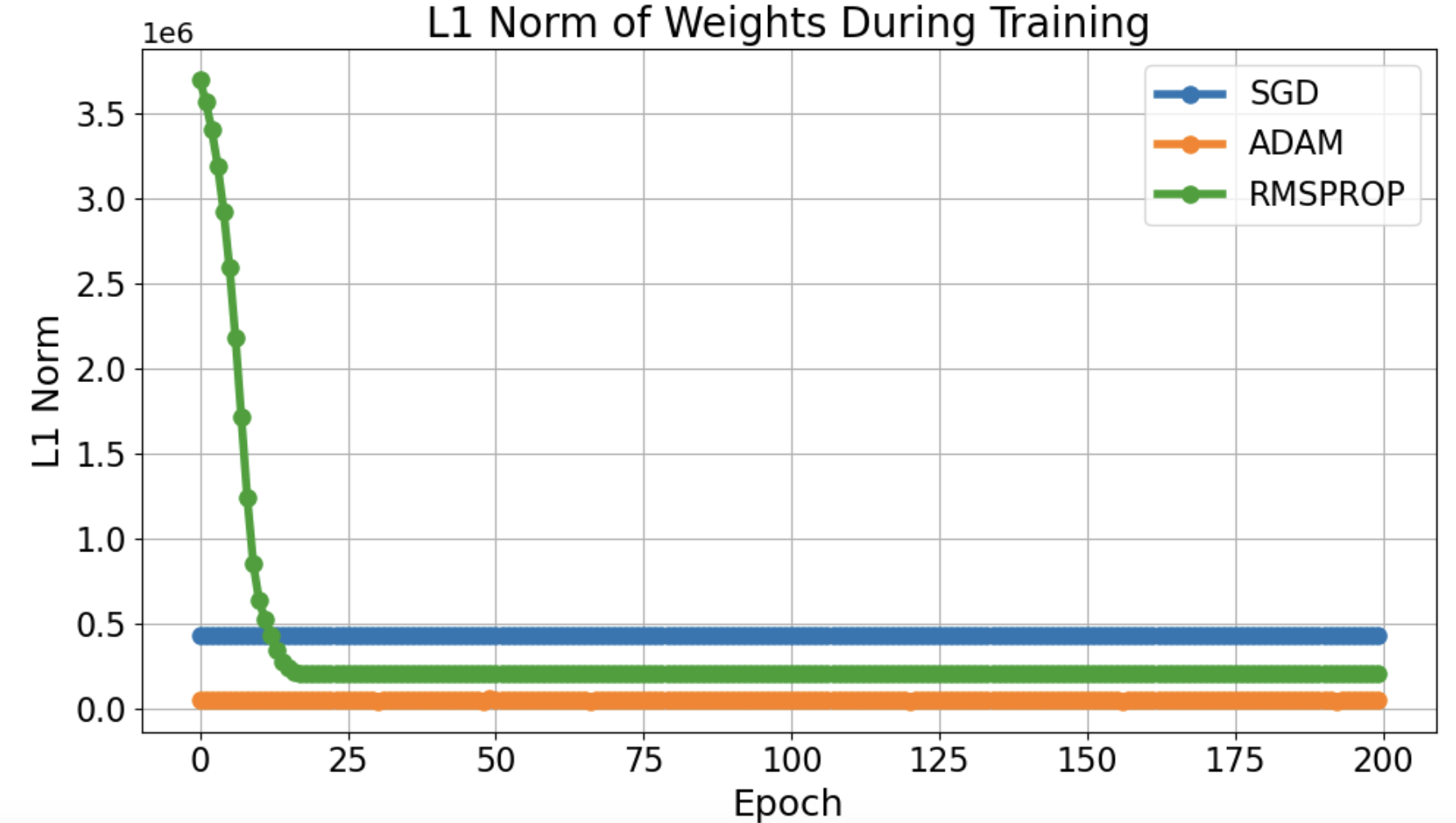}
  \vspace{0.5em}
\end{minipage}
\caption{$L_1$ norm of the weights as a function of iteration number for $\lambda_{1} = 0.002$ (left) and $\lambda_{1} = 2$ (right). In each figure, the green, orange, and blue lines represent curves for RMSProp, Adam, and vanilla SGD respectively.}
\label{fig:l1comparison}
\end{figure}

Our experiments yielded two key observations (Figure \ref{fig:l1comparison}): 1) The $L_1$ norms at convergence differ substantially across the three optimizer variants, and 2) for SGD and Adam, larger values of $\lambda_{1}$ result in a larger $L_1$ norm at convergence. These findings suggest that the relationship between $L_1$ regularization and prunable parameters may be more complex than previously understood. Indeed, our results indicate that a larger $\lambda_{1}$ does not necessarily correspond to more prunable parameters (Table \ref{tab:l1perccomp}): with $\lambda_{1} = 2$, less than 0.2\% of the parameters at convergence with vanilla NDGM have absolute value less than $10^{-5}$, but with $\lambda_{1} = 0.002$, more than 90\% do. Furthermore, these numbers are sensitive to the weight initialization and batch order during optimization bringing into question the notion of ``model interpretability" in the presence of stochasticity and non-differentiability. These observations raise important questions about the efficacy and generalizability of $L_1$ regularization for network pruning across different optimization algorithms and regularization strengths.

\paragraph{Soft-thresholding and ISTA:} The observation that higher $\lambda_{1}$ values do not yield more prunable parameters indicates that while soft thresholding produces sparse solutions for NDGMs, it does not find optimal solutions. This issue extends to general neural network problems, where any soft thresholding solution yields sparsity but not optimality. The root cause is that soft thresholding implicitly solves a generalized ISTA problem \citep{chen2023unified}, which assumes the loss function is $L-smooth$ (point 1, Section 3.2 of \citet{chen2023unified}) - an assumption that fails to capture the non-convex and non-differentiable dynamics in general neural network settings, as discussed previously. Consequently, applying the generalized ISTA formulation relying on $L-smoothness$ to neural networks necessarily leads to suboptimal solutions.

\paragraph{You Can’t Win with LASSO and NDGM:} $L_1$ regularization, widely used in deep learning, is expected to induce sparsity. The classical justification stems from its success in convex settings, particularly in feature selection and compressed sensing. However, the assumptions that make $L_1$ effective in these contexts break down in NDGMs, leading to unexpected and unreliable behavior.

In classical optimization and statistical settings, $L_1$ regularization is applied to loss functions to improve generalization. By trading off variance for some bias, it enhances the model’s performance on unseen data, and validation loss serves as a key measure to ensure proper model behavior. The optimization method used—whether coordinate descent or interior point methods in structured convex problems—is selected to guarantee the optimality of the solution. Moreover, convex solvers encapsulate the complexities associated with selecting stopping criteria and hyperparameters, requiring minimal user intervention for a given $\lambda_{1}$. Furthermore, the solution to this problem naturally exhibits sparsity, meaning there is no additional burden in enforcing it. This distinction is crucial: in frameworks like GLMNet or Scikit-learn, one typically counts only exact zeros to evaluate sparsity. Importantly, near-zero values are not manually thresholded, even in models such as Elastic Net, where the ridge penalty often results in several near-zero terms.

In deep learning, however, $L_1$ regularization is typically used for model compression, prioritizing a smaller model over generalization. Thus, the objective of $L_1$ regularization in deep learning is fundamentally different from its role in traditional settings. Furthermore, practitioners manually control both the algorithm (choosing gradient-based NDGMs) and the stopping criteria (deciding when training should end). The entire pipeline prioritizes flexibility over optimality. Given that NDGMs are designed for smooth problems, they are an ill-suited choice for solving optimization problems with sharp corners, such as those induced by $L_1$ regularization. As a result, it is unsurprising that NDGM solutions fail to exhibit true sparsity.

Typical training runs typically enforce an additional pruning step after $L_{1}$ regularization as is done in \cite{han2015deep} to compensate for the inability of NDGMs to generate sparse solutions. However this introduces an additional complication beyond NDGMs inability to generate sparse solutions -- it requires pruning, which is a form of soft thresholding, which will be suboptimal given our loss function is typically not L-smooth as discussed in our section on the ISTA.

\paragraph{NDGM and the way forward with sparsity:} While NDGMs do not inherently generate sparse solutions, iterative methods have been developed for $L_{1}$ regularized problems \citep{ma2019transformed,siegel2020training}. The realization that NDGMs lack this capability predates the widespread adoption of deep learning \citep{xiao2009dual}. Despite this, a misconception persists in the deep learning literature, suggesting that such algorithms yield sparse solutions comparable to those from traditional machine learning frameworks like glmnet \citep{friedman2021package} or scikit-learn \citep{kramer2016scikit}. However, there is a fundamental disparity between these frameworks: glmnet focuses on a narrow set of problems, guaranteeing optimality, while PyTorch \citep{paszke2017automatic} provides flexibility in model creation, but lacks similar guarantees. Attaining a solution at a non-differentiable point is challenging, emphasizing the importance of exploring related studies and tools rather than relying on simplistic implementations when precision is crucial.

\paragraph{Connections to the Edge of Stability:} Proposition \ref{prop:lassooverall} shows that the NDGM sequence for the non-differentiable, Lipschitz continuous, convex loss function described in \eref{las01} will not diverge to $\infty$ for any finite value of $\alpha$. This is a special case of unstable convergence, a topic we cover in greater detail in the next section.

\section{NDGM and the Edge of Stability} \label{sec:eos}

In recent years, there has been growing interest in the phenomenon of "unstable convergence" in neural network optimization. The literature on this subject \citep{ahn2022understanding,ahn2022learning,arora2022understanding,chen2022gradient,cohen2021gradient,li2022analyzing} is motivated by an intriguing observation: unlike convex quadratic forms, gradient descent on neural networks does not diverge to $\infty$ even when the learning rate exceeds $\alpha^{*} = 2/\eta$, where $\eta$ is the dominant eigenvalue of the loss function's Hessian. Instead, the loss function has been empirically shown to converge "unstably", reducing non-monotonically in the long run. The value $\alpha = \alpha^{*}$ is termed the "Edge of Stability" in the literature, as it demarcates regions of "stable convergence" ($\alpha < \alpha^{*}$) from regions of "unstable convergence" ($\alpha > \alpha^{*}$).

We define unstable convergence as "non-divergence to $\infty$". This definition is adapted from Definition 1.1 and the subsequent discussion in \citet{arora2022understanding}, which explains why using sharpness greater than $2/\eta$ is not part of the definition. We deviate from their definition due to its reliance on the Hessian existing at all points between the current and next iterate, a condition unmet in non-smooth neural networks. Importantly, the derivation of the Edge of Stability condition assumes that the loss function, $f$, is $L-smooth$ (see equation \eref{lsmooth}).

It has been conjectured that neural networks exhibit unstable convergence because this phenomenon lies beyond the scope of classical optimization theory \citep{cohen2021gradient,ahn2024learning}. However, we show that this conjecture is incorrect for a broader class of functions, specifically Lipschitz continuous, non-smooth convex functions satisfying \eref{lips}, as demonstrated in the following proposition:

\begin{proposition}\label{prop:eos}
    All convex non-smooth loss functions having bounded subgradients, or satisfying equation \eref{lips} will show unstable convergence.
\end{proposition}
\begin{proof}
    See Appendix \ref{app:eos} for details.
\end{proof}

The toy LASSO problem studied in \eref{las01} and the Huber loss function, widely used in robust regression and object detection models \citep{girshick2015fast,liu2016ssd,ren2015faster}, exemplify non-smooth convex loss functions satisfying Proposition \ref{prop:eos}. The LASSO problem is non-differentiable, while the Huber loss function is differentiable but non-smooth, as it is once but not twice differentiable. The Huber loss for regressing an arbitrary $50 \times 1$ response vector, on a \(50 \times 200\) data matrix is given by

\be
\begin{aligned}
L_7(\beta) &= \frac{1}{50} \sum l(i), \quad \text{where} \\
l(i) &=
\begin{cases}
    \frac{1}{2}(y_{i} - z_{i}^{T}\beta)^{2} & \text{if } \left | (y_{i} - z_{i}^{T}\beta) \right | < 1\\
    \left(|y_{i} - z_{i}^{T}\beta| - \frac{1}{2} \right) & \text{otherwise}
\end{cases}
\end{aligned}
\label{eq:huberloss}
\ee

\begin{figure}[!htbp]
\centering
\begin{minipage}{0.48\textwidth}
  \centering
  \includegraphics[width=0.48\textwidth]{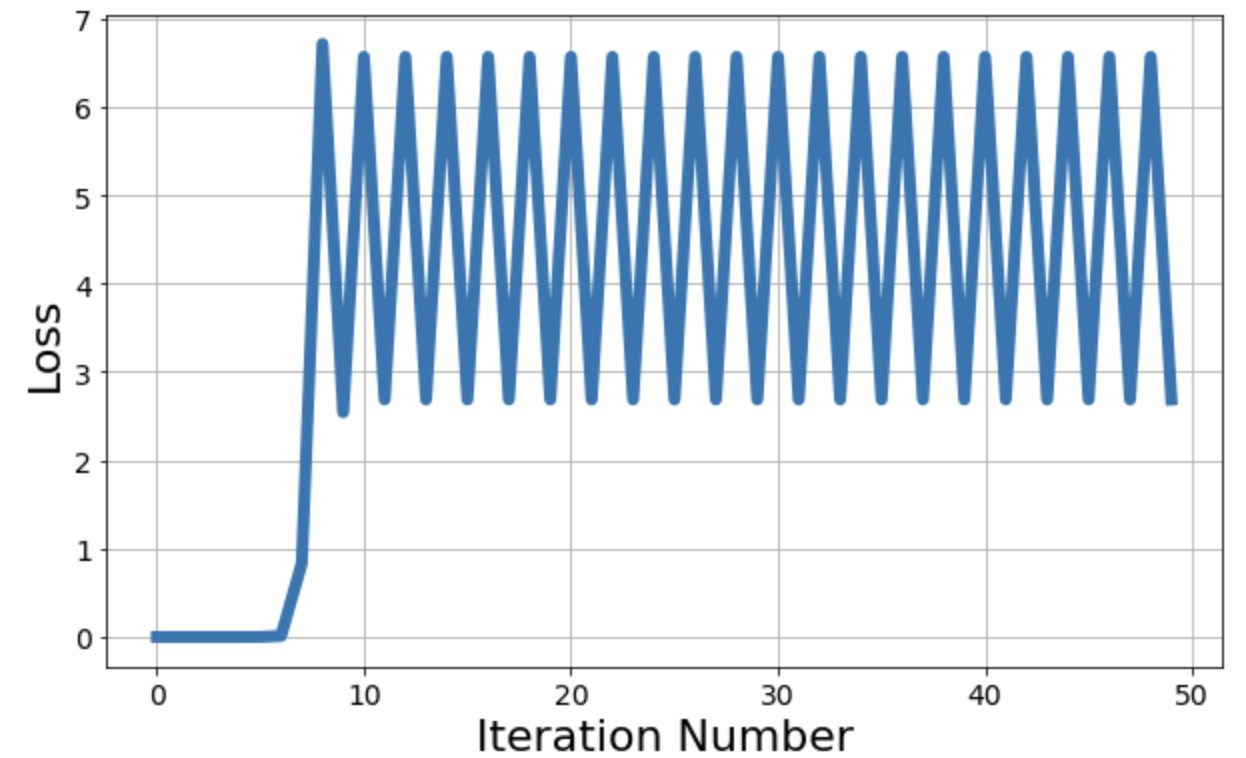}\hfill
  \includegraphics[width=0.44\textwidth]{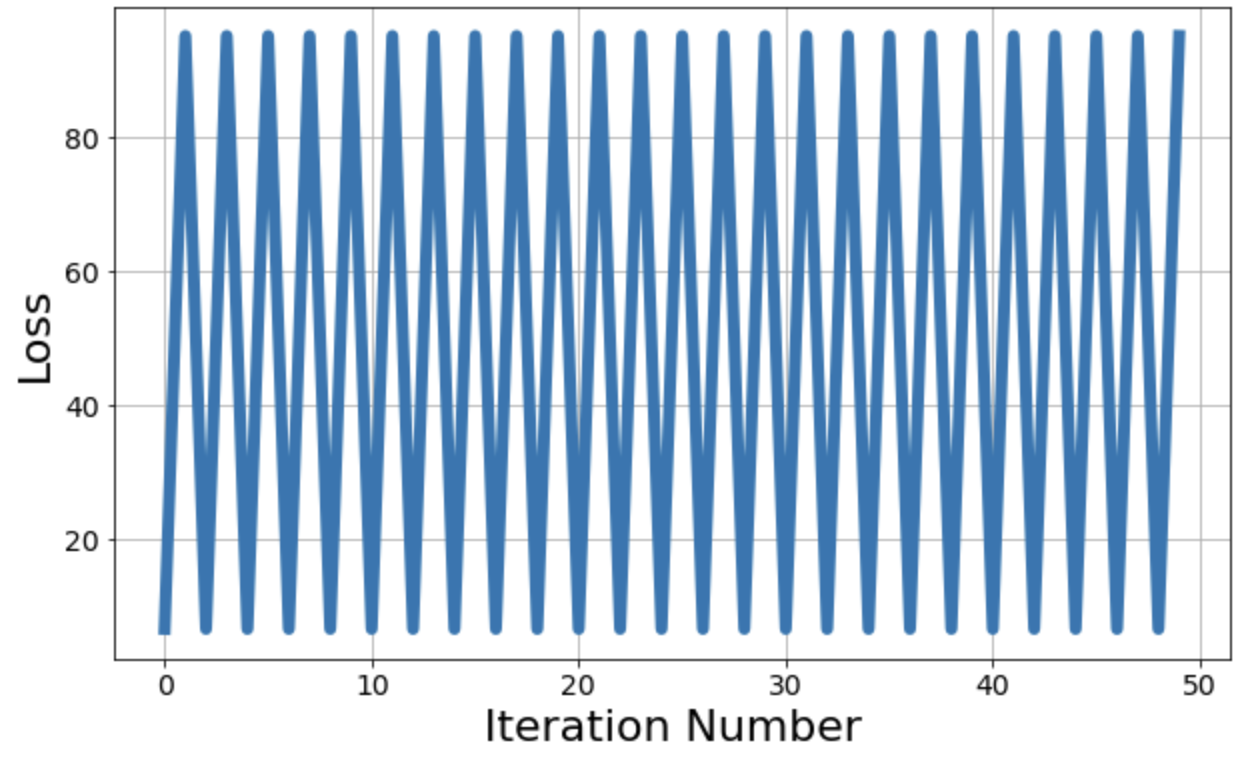}
  \vspace{0.5em}
\end{minipage}
\caption{Training curves for $\alpha = 0.1$ and $\alpha = 10$ with the Lipschitz continuous Huber loss function. As the learning rate increases, the loss starts oscillating with larger frequencies, but never diverges to $\infty$. This is to be expected from Proposition \ref{prop:eos}, since the Huber loss function is convex, non-smooth and has bounded subgradients.}
\label{fig:huberloss}
\end{figure}

Figure \ref{fig:huberloss} illustrates that the NDGM sequence for \eref{huberloss} does not diverge toward $\infty$, even with a high learning rate such as $\alpha = 10$, aligning with our expectations.

\paragraph{Subquadratic losses and unstable convergence:} In section 3.1 of \citet{ma2022multiscale}, the authors use second-order finite differences to analyze the curvature of VGG16 and ResNet near local minima, concluding that subquadratic growth explains the edge of stability phenomenon. However, their analysis is fundamentally flawed for these ReLU networks because second-order finite differences break down at their generically non-differentiable local minima \citep{laurent2018multilinear}. The overparameterized Huber loss function (Equation \ref{eq:huberloss}) further challenges the general applicability of their claims: despite having a global minimum of 0 with quadratic growth nearby, it exhibits unstable convergence as guaranteed by Proposition \ref{prop:eos} (Figure \ref{fig:huberloss}). This discrepancy arises because the Huber loss lacks the twice differentiability required by Definition 4.1 in \citet{ma2022multiscale}. In contrast, all $L-smooth$ functions, being twice differentiable by definition, will show subquadratic growth near their local minima during unstable convergence, aligning with their results. This example underscores the risks of extending results derived for smooth functions to non-smooth cases.

\paragraph{The way forward:} The limitations of applying results like those in \citet{cohen2021gradient} to non-smooth settings highlight the need for new theoretical frameworks. In hindsight, it is not particularly surprising that \citet{cohen2021gradient}.'s results do not hold even for the differentiable case, as they derive their key result using the assumption of $L-smoothness$ while acknowledging in their discussion that this assumption is not ideal for neural networks. We propose using the convex non-differentiable single-layer neural network from \citet{kumar2023analytical} as a toy problem for theory development. This approach offers advantages in terms of analytical tractability and relevance to ReLU networks, potentially leading to insights about optimization around non-differentiable points and informing more robust deep learning optimization techniques.

\section{Painful Pathologies of Poorly Programmed Perturbations}  Training of modern LLMs often follows schedules derived from empirical scaling laws \citep{kaplan2020scaling} and compute-optimal insights such as those in \citet{hoffmann2022training}. These protocols typically advocate stopping well before full convergence—arguing that beyond a certain point, the returns diminish while compute costs dominate. The prevailing view is that compute, not convergence, is the primary bottleneck, and that training all the way to the global minimum is neither feasible nor necessary.

But even if the minimum remains out of reach, non-differentiability cannot be ignored. The breakdown of differentiability—and with it, the erosion of duality and geometric support—is not an asymptotic fluke. It is a structural inevitability in wide, deep, non-smooth architectures. As training descends deeper into the loss landscape, the geometry turns hostile: support vanishes, non-verticality collapses, and oscillations emerge—not due to bad hyperparameters, but because the landscape itself no longer admits reliable directional information.

This makes it imperative to analyze the dynamics of modern training under non-differentiability and finite step sizes—not as pathological edge cases, but as central features of the large-scale regime.

\section{Ethics and AI publishing}

There is a growing disconnect between what is analyzable and what is deployed—and it is no longer sufficient to dismiss this as an artifact of a "young field" or "practical constraints." When theoretical claims are built on assumptions known to be violated in the very systems they aim to explain, and these claims are routinely published, cited, and used to shape research agendas, the issue is not just technical—it is ethical. Not in the narrow sense of harm or fairness, but in the deeper sense of intellectual stewardship. As a community, we must hold ourselves to higher standards of conceptual clarity and accountability. Simplifying models to make them analyzable is not inherently flawed—but presenting those analyses as representative without qualifying the mismatch is. We cannot keep watering down theorems until they lose contact with reality, nor can we afford to confuse convenience for insight. Without course correction, we risk turning a foundational science into a scaffolding of performance-chasing approximations. The burden is on us—not reviewers, not citation counts—to keep the building intact.

\section{Conclusion}
This paper demonstrates how non-differentiability fundamentally alters neural network training dynamics, challenging widely held assumptions in deep learning optimization. Our findings in convergence analysis, LASSO problems, and the Edge of Stability phenomenon reveal that many current practices based on smooth optimization theory can lead to suboptimal or incorrect results in non-smooth settings. For the deep learning community, these results underscore the need to reevaluate common techniques in network pruning, learning rate selection, and convergence analysis. We call for a collective effort to develop new theoretical frameworks and empirical validations that accurately capture the behavior of non-smooth optimization in deep learning. By building on more appropriate foundations, we can potentially unlock more efficient and reliable training algorithms, leading to improved performance across a wide range of applications. As the field continues to evolve, it is crucial that both researchers and practitioners approach optimization in neural networks with a more nuanced understanding of non-smoothness and its implications.

\section{Acknowledgements}
We would like to thank David Steinsaltz, Kenneth Wachter, Wenyun Zuo, Ryan Tibshirani, Rob Tibshirani, Stephen Boyd, Trevor Hastie, Marc Feldman, Shripad Tuljapurkar and Boris Ginsbourg for their valuable feedback on early drafts of this paper. We are particularly indebted to Dmitri Bertsekas whose excellent books fostered our interest in this subject. Finally, we thank our PhD advisors, Marc Feldman and Shripad Tuljapurkar, who taught us how to think about dynamical systems.

\bibliographystyle{plainnat}
\bibliography{aistats2k25}

\clearpage
\onecolumn
\appendix

\section*{SUPPLEMENTARY MATERIAL}

\section{Appendices for section \ref{sec:ndgm-convergence}} \label{app:convanalysis}

\subsection{The Capture Theorem and its nuances}
The statement of the Capture Theorem (Proposition 1.2.5 from \cite{bertsekas1997nonlinear}), is as follows:

\begin{proposition}
\textbf{(Capture Theorem)} Let $f$ be continuously differentiable, and let $\{x_k\}$ be a sequence satisfying $f(x_{k+1}) \leq f(x_k)$ for all $k$, generated by a gradient method $x_{k+1} = x_k + \alpha_k d_k$, where every limit point of the sequence is a stationary point of $f$. Suppose there exist constants $s > 0$ and $c > 0$ such that for all $k$, $\alpha_k \leq s$, and $\|d_k\| \leq c \|\nabla f(x_k)\|$. Let $x^*$ be a local minimum of $f$, and assume it is the only stationary point in some open neighborhood. Then there exists an open set $S$ containing $x^*$ such that if $x_p \in S$, then $x_q \in S$ for all $q \geq p$. Furthermore, for any $\epsilon > 0$, the set $S$ can be chosen such that $\|x - x^*\| < \epsilon$ for all $x \in S$.
\end{proposition}

\vspace{1em}
We unpack the implications of this result and contrast the expected behavior under the theorem with the behavior of NDGM in a non-smooth setting.

\subsubsection{Satisfiability under a general norm}

Neither the statement of the Capture Theorem provided above nor its proof  in \cite{bertsekas1997nonlinear} depends on the use of the 2-norm. Both rely only on the non-negativity of norms and the triangle inequality, and are therefore valid for all norms. In our experiments illustrating the Capture Theorem, we use the 1-norm, but the reader may verify that the result holds under other norms as well. To emphasize this, we provide an explicit example of the capturing behavior using the infinity norm in the discussion below.

\subsubsection{Monotonicity Need Not Start at Iteration 0}

A common misinterpretation of the Capture Theorem is that the norm of iterates must decrease monotonically from the start of optimization. This is not required. The theorem only guarantees that \emph{once the sequence enters a neighborhood of a local minimum}, the iterates remain captured within it. That is, there exists an $M$ such that for all $k \geq M$, the trajectory remains in the capture region.

Since gradient descent generates an infinite sequence, one can always relabel a monotone subsequence starting at $x_M$ as $x_0$, which may give the misleading impression that convergence behavior (in norm) should begin immediately.

\begin{figure}[h]
    \centering
    \includegraphics[width=0.6\textwidth]{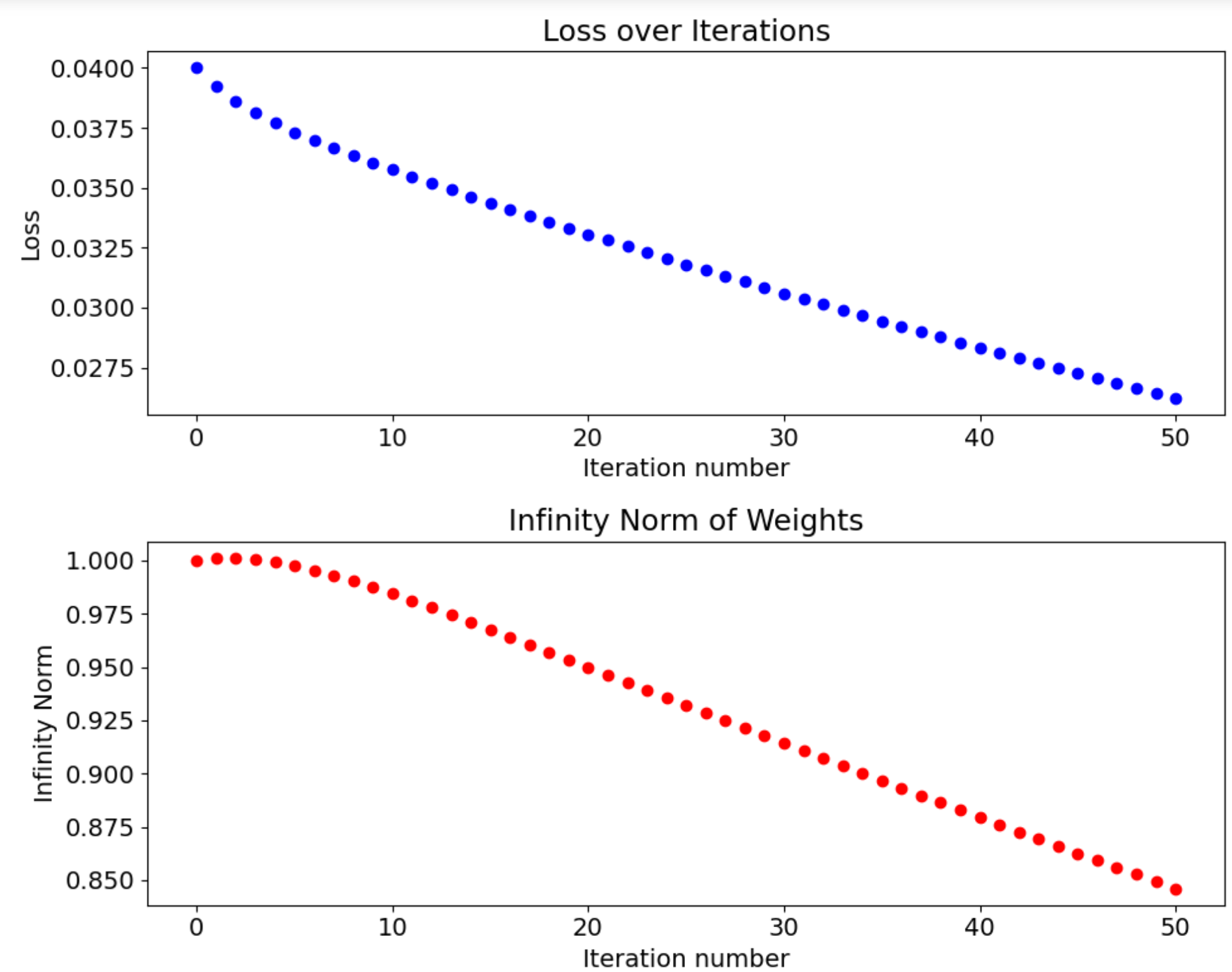}
    \caption{Infinity norm of weights during gradient descent on a smooth quadratic problem. Initial steps increase in norm, but the trajectory eventually stabilizes, consistent with the Capture Theorem.}
    \label{fig:qp_plot_inf_norm}
\end{figure}

To illustrate this, we consider gradient descent applied to a quadratic form:
\[
f(x) = \frac{1}{2} x^T H x \quad \text{where} \quad H = \begin{bmatrix} 1 & -0.91 \\ -0.91 & 0.9 \end{bmatrix}.
\]
This function is convex and $L$-smooth, with dominant eigenvalue $\lambda_1 \approx 1.86$. Using a constant step size of $\alpha = 1 / \lambda_1$, gradient descent is guaranteed to converge. As shown in Figure~\ref{fig:qp_plot_inf_norm}, the $L_\infty$ norm of $x_k$ initially increases, but then begins to decay monotonically, in line with the Capture Theorem's predictions.\footnote{We thank an anonymous reviewer for providing us with this example}

\subsubsection{NDGM Behavior: Monotonic Loss Without Capture}

We contrast the smooth case above with a ReLU-based regression problem optimized via a subgradient method (NDGM) using adaptive step sizes. In this experiment, whenever the loss increases, the step size is reduced multiplicatively. This enforcement ensures a \emph{monotone, bounded} loss sequence. Since the loss is always non-negative, it converges (in the sense of real analysis), even though it does not converge to zero due to the finite run.

Despite the monotone loss, we observe that the weight iterates \(\{\beta_k\}\) \emph{never stabilize}. As shown in Figure~\ref{fig:relu_plot_one_norm}, the $L_1$ norm oscillates indefinitely within two distinct bands. Notably, this oscillation is not a byproduct of finite precision: the model is implemented in double precision, and the bands remain separated by a gap on the order of \(10^{-14}\). Furthermore, the loss for the last 1000 iterations is identically 4461.786221675728, so the loss is indeed monotone non-increasing. The behavior is real, not numerical. This is a clear violation of the capture behavior one expects from gradient descent. Importantly, this behavior is not a function of the step size (which is being controlled), nor of the norm (as the Capture Theorem holds for any norm). The breakdown arises solely from the fact that the objective is \emph{not differentiable}.

\subsubsection{Conclusion}

This experiment highlights a fundamental bifurcation in behavior: under smoothness, monotonic loss implies eventual stabilization of the iterates (as captured by the Capture Theorem). Under non-smoothness, this implication fails. The optimizer can “descend forever” in function value while continuing to oscillate in parameter space. This dynamic is structurally similar to Edge of Stability behavior observed in deep networks, and further underscores the need for non-smooth theory to guide our understanding of training dynamics in modern architectures.

\subsubsection{Code to generate the plot}

The code is GPT generated and validated for correctness
\begin{verbatim}

import torch
import torch.nn as nn
import numpy as np
import matplotlib.pyplot as plt
import matplotlib
from torch.utils.data import TensorDataset, DataLoader
import torch.optim as optim

# Set random seed for reproducibility
seed = 42
torch.manual_seed(seed)
np.random.seed(seed)

# --- Data Generation ---
def generate_dummy_data(x_shape=(20, 50000), x_range=(-1, 1), y_range=(-1, 1)):
    """
    Generates dummy training data.
    
    Args:
        x_shape (tuple): Shape of the input features (n_samples, n_features).
        x_range (tuple): Range of values for input features.
        y_range (tuple): Range of values for response vector.
    
    Returns:
        Tuple[Tensor, Tensor]: x_train and y_train tensors.
    """
    x_train = np.random.uniform(*x_range, size=x_shape)
    y_train = np.random.uniform(*y_range, size=(x_shape[0], 1))
    return torch.Tensor(x_train), torch.Tensor(y_train)

# --- Model Definitions ---
class BaseNN(nn.Module):
    """
    Single-layer ReLU neural network.
    """
    def __init__(self, input_size, init_low=-0.99, init_high=0.99):
        super(BaseNN, self).__init__()
        self.linear = nn.Linear(input_size, 1, bias=False).double()
        torch.nn.init.uniform_(self.linear.weight, init_low, init_high)
        self.relu = nn.ReLU()

    def forward(self, x):
        return self.relu(self.linear(x))

class LinearReg(nn.Module):
    """
    Basic linear regression model.
    """
    def __init__(self, input_size, init_low=-0.99, init_high=0.99):
        super(LinearReg, self).__init__()
        self.linear = nn.Linear(input_size, 1, bias=False)
        torch.nn.init.uniform_(self.linear.weight, init_low, init_high)

    def forward(self, x):
        return self.linear(x)

# --- Training Function ---
def runner(model, x, y, lr=0.1, max_iter=5000, l2_penalty=0.0, l1_penalty=0.0, 
           criterion=nn.MSELoss(reduction='sum'), patience=100, tol=1e-8, decay_factor=0.9):
    """
    SGD with adaptive LR:
      - decay immediately on loss increase
      - decay if loss is flat for `patience` iterations
    """
    optimizer = optim.SGD(model.parameters(), lr=lr)
    losses, weights, lrs, inf_norms = [], [], [], []
    recent_losses = []
    prev_loss = float('inf')

    for it in range(max_iter):
        out = model(x)
        loss = criterion(out, y)
        wt_sqrd = model.linear.weight.pow(2).sum()
        l2_norm = l2_penalty * wt_sqrd
        l1_norm = l1_penalty * torch.norm(model.linear.weight, p=1)
        loss += l2_norm + l1_norm

        curr_loss = loss.item()
        losses.append(curr_loss)
        weights.append(np.copy(model.linear.weight.detach().numpy()))
        inf_norms.append(model.linear.weight.abs().max().item())
        lrs.append(optimizer.param_groups[0]['lr'])

        # ---- Adaptive LR triggers ----

        # 1. Immediate decay on loss increase
        if curr_loss > prev_loss:
            for g in optimizer.param_groups:
                g['lr'] *= decay_factor
            recent_losses.clear()
        else:
            recent_losses.append(curr_loss)
            if len(recent_losses) > patience:
                recent_losses.pop(0)
                if curr_loss > prev_loss:
                    for g in optimizer.param_groups:
                        g['lr'] *= decay_factor
                    recent_losses.clear()

        prev_loss = curr_loss

        loss.backward()
        optimizer.step()
        optimizer.zero_grad()

    return losses, weights

# --- Experiment Setup ---
nrows = 5000
ncols = 500
l1 = 0.001
l2 = 0

x, y = generate_dummy_data(x_shape=(nrows, ncols), x_range=(-1, 1), y_range=(-0.99, -0.9))
model = BaseNN(input_size=ncols, init_low=-1, init_high=1)

losses, weights = runner(model=model, x=x.double(),
y=y.double(), lr=0.01, l1_penalty=l1, l2_penalty=l2, max_iter=6000)
one_norm = [np.linalg.norm(w[0], 1) for w in weights]
ticks = list(range(len(losses)))

# --- Visualization ---
matplotlib.rcParams.update({'font.size': 14})
fig, (ax1, ax2) = plt.subplots(2, figsize=(10, 8))


ax1.set_yscale('log')
ax1.plot(ticks[-1000:], losses[-1000:], 'bo')
ax1.set(xlabel="Iteration number", ylabel="Loss", title="Loss over last 1000 Iterations")

ax2.plot(ticks[-1000:], one_norm[-1000:], 'ro')
ax2.set(xlabel="Iteration number", ylabel="One Norm", title="One Norm over last 1000 iterations")

plt.tight_layout()
plt.show()

\end{verbatim}

\subsection{Code for activation function comparison experiments} \label{app:relugelucomp}
The code is GPT generated and validated for correctness

\begin{verbatim}
import torch
import torch.nn as nn
import torch.optim as optim
import torchvision
from torchvision import transforms
from torch.utils.data import DataLoader
import matplotlib.pyplot as plt
from collections import defaultdict
import random
import numpy as np

# -----------------------
# Set Global Seed
# -----------------------
def set_seed(seed=42):
    torch.manual_seed(seed)
    torch.cuda.manual_seed_all(seed)
    np.random.seed(seed)
    random.seed(seed)
    torch.backends.cudnn.deterministic = True
    torch.backends.cudnn.benchmark = False

set_seed()

# -----------------------
# Device
# -----------------------
device = torch.device("cuda" if torch.cuda.is_available() else "cpu")
print("Device:", device)

# -----------------------
# Data Transforms
# -----------------------
transform = transforms.Compose([
    transforms.RandomCrop(32, padding=4),
    transforms.RandomHorizontalFlip(),
    transforms.ToTensor(),
    transforms.Normalize((0.4914, 0.4822, 0.4465), (0.2023, 0.1994, 0.2010)),
])

trainset = torchvision.datasets.CIFAR10(root='./data', train=True, download=True, transform=transform)
testset = torchvision.datasets.CIFAR10(root='./data', train=False, download=True, transform=transform)

trainloader = DataLoader(trainset, batch_size=128, shuffle=True, num_workers=2)
testloader = DataLoader(testset, batch_size=128, shuffle=False, num_workers=2)

# -----------------------
# VGG Config
# -----------------------
cfg = {
    'VGG11': [16, 'M', 32, 'M', 32, 32, 'M', 64, 64, 'M', 64, 64, 'M'],
    'VGG13': [16, 16, 'M', 32, 32, 'M', 32, 32, 'M', 64, 64, 'M', 64, 64, 'M'],
    'VGG16': [16, 16, 'M', 32, 32, 'M', 32, 32, 32, 'M', 64, 64, 64, 'M', 64, 64, 64, 'M'],
    'VGG19': [16, 16, 'M', 32, 32, 'M', 32, 32, 32, 32, 'M', 64, 64, 64, 64, 'M', 64, 64, 64, 64, 'M'],
}


# -----------------------
# VGG Model (modular)
# -----------------------
class VGG(nn.Module):
    def __init__(self, model_name, activation_type='relu', softplus_beta=1.0, include_bn_layer=False):
        super(VGG, self).__init__()
        self.activation_type = activation_type.lower()
        self.softplus_beta = softplus_beta
        self.include_bn_layer = include_bn_layer
        self.features = self._make_layers(cfg[model_name])
        self.classifier = nn.Linear(64, 10)

    def forward(self, x):
        x = self.features(x)
        x = x.view(x.size(0), -1)
        return self.classifier(x)

    def get_activation(self):
        if self.activation_type == 'relu':
            return nn.ReLU(inplace=True)
        elif self.activation_type == 'gelu':
            return nn.GELU()
        elif self.activation_type == 'softplus':
            return nn.Softplus(beta=self.softplus_beta)
        else:
            raise ValueError(f"Unsupported activation: {self.activation_type}")

    def _make_layers(self, config):
        layers = []
        in_channels = 3
        activation = self.get_activation()

        for x in config:
            if x == 'M':
                layers.append(nn.AvgPool2d(kernel_size=2, stride=2))
            else:
                layers.append(nn.Conv2d(in_channels, x, kernel_size=3, padding=1))
                if self.include_bn_layer:
                    layers.append(nn.BatchNorm2d(x))
                layers.append(activation)
                in_channels = x

        layers.append(nn.AvgPool2d(kernel_size=1, stride=1))
        return nn.Sequential(*layers)

# -----------------------
# Compute Validation Loss
# -----------------------
def compute_validation_loss(model):
    model.eval()
    val_loss, total = 0.0, 0
    criterion = nn.CrossEntropyLoss()
    with torch.no_grad():
        for inputs, targets in testloader:
            inputs, targets = inputs.to(device), targets.to(device)
            outputs = model(inputs)
            loss = criterion(outputs, targets)
            val_loss += loss.item()
            total += targets.size(0)
    return val_loss / len(testloader)

# -----------------------
# Main Training Function
# -----------------------
def train_and_compare(model_id, num_epochs=100, include_bn=False):
    activation_configs = [
        ('relu', None),
        ('gelu', None),
        ('softplus', 100.0),
        ('softplus', 1000.0)
    ]

    histories = defaultdict(list)

    for act, beta in activation_configs:
        print(f"\nTraining {model_id} with {act.upper()} (beta={beta})")

        set_seed(19)
        model = VGG(model_id, activation_type=act, 
        softplus_beta=beta or 1.0, include_bn_layer=include_bn).to(device)
        optimizer = optim.Adam(model.parameters(), lr=0.001)
        criterion = nn.CrossEntropyLoss()

        for epoch in range(num_epochs):
            model.train()
            epoch_loss = 0.0

            for inputs, targets in trainloader:
                inputs, targets = inputs.to(device), targets.to(device)
                optimizer.zero_grad()
                outputs = model(inputs)
                loss = criterion(outputs, targets)
                loss.backward()
                optimizer.step()
                epoch_loss += loss.item()

            avg_loss = epoch_loss / len(trainloader)
            val_loss = compute_validation_loss(model)

            print(f"Epoch {epoch:03d} | Train Loss: {avg_loss:.4f} | Val Loss: {val_loss:.4f}")
            act_key = f'{act}_{beta}' if beta is not None else act
            histories[(act_key, 'train')].append(avg_loss)
            histories[(act_key, 'val')].append(val_loss)

    return histories

results_vgg11 = train_and_compare('VGG11', num_epochs=20, include_bn=False)
results_vgg13 = train_and_compare('VGG13', num_epochs=20, include_bn=False)
results_vgg16 = train_and_compare('VGG16', num_epochs=20, include_bn=False)
results_vgg19 = train_and_compare('VGG19', num_epochs=20, include_bn=False)


results_vgg11_bn = train_and_compare('VGG11', num_epochs=20, include_bn=True)
results_vgg13_bn = train_and_compare('VGG13', num_epochs=20, include_bn=True)
results_vgg16_bn = train_and_compare('VGG16', num_epochs=20, include_bn=True)
results_vgg19_bn = train_and_compare('VGG19', num_epochs=20, include_bn=True)
\end{verbatim}

\subsection{Dense Digressions into Devious Duality}\label{app:cvxanal}
The loss function in \ref{eq:prel0} exhibits strong duality, which is typically established using Slater's condition since 0 is an interior point of the domain. In this subsection, we provide a geometric characterization of this fact. Interestingly, these geometric properties extend beyond our specific case to more general settings. We present these general results before specializing to our setup.

\subsubsection{Preliminaries and Notation}

Throughout our analysis, we work with extended value convex functions $f: \mathbb{R}^n \rightarrow \mathbb{R} \cup \{-\infty, \infty\}$. This formulation allows us to work directly in $\mathbb{R}^n$, as constraints are automatically handled through the extended value representation: points outside the feasible domain simply take infinite values. A fundamental object in our study is the epigraph of $f$, defined as $M = \text{Epi}(f) = \{(x, w) : f(x) \leq w\}$. A key property we utilize is that the epigraph of a convex function is itself a convex set (Section 3.1.7 in \cite{boyd2004convex}), providing the geometric foundation for our subsequent analysis.
\subsubsection{Geometric Properties of Global Minima} 

We show that there exists a non-empty family of non-vertical hyperplanes passing through the point on the epigraph corresponding to the global minimum of $f$. Moreover, no other hyperplane containing the epigraph can have a higher intercept on the $(n+1)$st axis.

To establish this result, we first analyze the case where $f$ has a global minimum at 0. The Min Common Point/Max Crossing (MC/MC) framework (for details, see Chapter 4 of \cite{bertsekas2009convex}) provides the tools for our analysis:

\begin{itemize}
  \item Min Common point ($w$*): The point with minimum $(n+1)$th component among all vectors common to $M$ and the $(n+1)$st axis. 
  \item Max Crossing ($q$*): The highest crossing point on the $(n+1)$st axis among all non-vertical hyperplanes containing $M$ in their upper half-space.
\end{itemize}

In our specific case, since $f$ has a global minimum at 0, $w^{*}$ represents the global lowest point on the epigraph and represents the lowest point of the epigraph along the $(n+1)$st axis. Our proofs utilize the following result:

\begin{proposition}\label{prop:bertlsc}
Let $M = \text{Epi}(p)$, where $p: \mathbb{R}^n \rightarrow [-\infty, \infty]$ is a convex function with $p(0) = w^* < \infty$. Then $q^* = w^*$ if and only if $p$ is lower semicontinuous at 0.
\end{proposition}
\begin{proof}
  See corollary to Proposition 4.3.1 in \cite{bertsekas2009convex} (Page 148).
\end{proof}

Using this result, we establish strong duality for the case when $f$ has a finite global minimum at 0:

\begin{proposition}\label{prop:lsc}
For $f$ as above, $q^* = w^* = f(0)$.
\end{proposition}
\begin{proof}
Let $g_k = f(u_k)$ for any sequence $u_k \rightarrow 0$. Since 0 is the global minimum of $f$, $g_k \geq f(0)$. Taking the limit,
\[
\liminf_{k \rightarrow \infty} g_k \geq f(0) = w^*,
\]
establishing lower semicontinuity at 0. The result follows from Proposition \ref{prop:bertlsc}.
\end{proof}

The equality $q^* = w^*$ suggests special geometric properties at $(0,\ldots,0,w^*)$. This point, where the epigraph meets the $(n+1)$st axis at its lowest value, warrants closer examination.
\begin{proposition}\label{prop:hyperplane}
There exists a non-empty family of non-vertical hyperplanes $\mathcal{H}$ passing through $(0,\ldots,0,f(0))$ that contain $M$ in their upper half-space. No other hyperplane containing $M$ in its upper half-space can have a higher intercept on the $(n+1)$st axis.
\end{proposition}

\begin{proof}
We are aware of four ways of establishing this result: one geometrical, and three analytical. The first two approaches rely on the fact that $(0,\ldots,0,w^*)$ is not an interior point of $M$, so we begin by establishing this claim before presenting the main argument from all three perspectives.

To show that $(0,\ldots,0,w^*)$ is not an interior point of $M$, suppose by contradiction that it is.
Then there exists $\epsilon > 0$ such that $B_{\epsilon}((0,\ldots,0,w^*)) \subset M$. This implies $(0,\ldots,0,w^*-\epsilon/2) \in M$, contradicting that $w^*$ is the value at the global minimum. 

\paragraph{The geometric approach: }By the supporting hyperplane theorem, there exists a supporting hyperplane passing through $(0,\ldots,0,w^*)$, and therefore, $\mathcal{H}$ is non-empty. These hyperplanes necessarily intersect the $(n+1)$st axis at $w^*$. By weak duality (Proposition 4.1.2 in \cite{bertsekas2009convex}), $q^* \leq w^*$, implying these hyperplanes achieve the highest possible intercept. 

To prove non-verticality, suppose by contradiction that some $h \in \mathcal{H}$ is vertical. Since $h$ passes through $(0,\ldots,0,w^*)$, it must have the form $q^T\beta = 0$ for some $q \in \mathbb{R}^n$, where $q \neq 0$. Construct $u \in \mathbb{R}^n$ by setting $u[i] = \text{sign}(q[i])$ for each component. Then $q^Tu > 0$ by construction. Note that points $(u, f(u))$ and $(-u, f(-u))$ must belong to $M$, where $f(u)$ and $f(-u)$ represent the height of the epigraph at these points. The heights at these points can be $\infty$ but not $-\infty$ as this would violate the existence of a finite global minimum. Importantly, the actual values of these heights do not affect our argument. Since $M$ must lie in the upper half-space of $h$, we require $q^T(-u) > 0$. However, $q^T(-u) = -(q^Tu) < 0$, a contradiction.

\paragraph{Analytical approach 1 - Using the non-vertical hyperplane theorem: } Since 1) $(0,\ldots,0,w^*)$ is not in the interior of $M$, and 2) $M$ is a convex set with no vertical lines,  the non-vertical hyperplane theorem (Proposition 1.5.8 in \citet{bertsekas2009convex}) states that there must exist a non-vertical hyperplane passing through $(0,\ldots,0,w^*)$ that contains $M$ in its positive halfspace. Furthermore, by the definition of max-crossing, the highest intercept on the $(n+1)st$ axis of any hyperplane containing $M$ in its positive halfspace is $q^{*}$. Therefore, for any $(u,w) \in M$, there exists some $\mu \in R^{N}$, such that
\be
\mu^{T} u + w \geq q^{*} \geq w^{*}. 
\ee 
However, by weak duality (Proposition 4.1.2 in \citet{bertsekas2009convex}) we have $w^{*} \geq q^{*}$, so combining the two inequalities gives $q^{*} = w^{*}.$

\paragraph{Analytical approach 2 - Using the non-vertical hyperplane theorem and the sandwich theorem:} To derive the result a third way, we first show that $(0,w^{*} - \epsilon) \notin cl(M)$ for any $\epsilon > 0$. To see why, suppose by contradiction that it is. Then, there would exist a sequence $(u_k, w_k)\in M$ such that $u_k \rightarrow 0$ and $w_k \rightarrow w^{*} - \epsilon$, which would imply that there exists an $N_0$ such that $w_{N} < w^{*}$ $\forall  N \geq N_{0}$. This is not possible since $w^{*}$ is the global minimum, thus establishing the contradiction. 

Accordingly, by the non-vertical hyperplane theorem (Proposition 1.5.8b in \citet{bertsekas2009convex}), there exists a non vertical hyperplane that strictly separates $(0, w^{*} - \epsilon)$ and $M$. Consider such a hyperplane which has an intercept of $\xi$ on the $(n+1)^{st}$ axis, and contains $(0,w^{*} - \epsilon)$ in the negative halfspace, and $M$ in the positive halfspace. Since $q^{*}$ is the highest intercept of a non-vertical hyperplane having $M$ in its positive halfspace, we have $\xi \leq q^{*}$, and by weak duality we have $q^{*} \leq w^{*}$, so putting it altogether gives 
\be
w^{*} - \epsilon < \xi \leq q^{*} \leq w^{*}.
\ee 
The result follows as $\epsilon \rightarrow 0$.

\paragraph{Analytical approach 3 - Using the MC/MC Strong Duality theorem:} This is essentially a higher abstraction of Analytical Approach 2. Indeed, the sandwich theorem technique employed in our previous proof is taken near verbatim from the proof of the MC/MC Strong Duality theorem (Proposition 4.3.1 in \citet{bertsekas2009convex}). For the purposes of our result, the theorem states that if 1) $w^{*} < \infty$, 2) $\overline{M} = M + \left\{(0,w) | w \geq 0 \right\}$ is convex, and 3) if the function is lower semi-continuous at 0, then $q^{*} = w^{*}$. The first two points are trivially established (see Proposition \ref{prop:compactness} for details if needed), and the third point is established in the proof of Proposition \ref{prop:lsc}, and therefore we have $q^{*} = w^{*}$. This result does not come as a surprise, since Proposition \ref{prop:bertlsc} is a corollary of the MC/MC strong duality theorem, thus closing the loop on the above propositions.
\end{proof}
\paragraph{What a difference finiteness makes!:} The result above has strong connections with the non-vertical hyperplane theorem stated below for completeness:

\begin{proposition}
    Let $C$ be a non empty convex subset of $R^{N+1}$ that contains no vertical lines. Let the vector in $R^{N+1}$ be denoted by $(u,w)$, where $u \in R^{N}$ and $w \in R$. Then $C$ is contained in a closed halfspace corresponding to a non-vertical hyperplane.
\end{proposition}
\begin{proof}
    Simplified version of Proposition 1.5.8 in \citet{bertsekas2009convex}.
\end{proof}

Note that the above proposition is non-comittal on whether the non-vertical hyperplane ``touches" $C$ or not. Our proposition shows that ``touching" is guaranteed if the global minimum is finite, thus providing the bridge between weak and strong duality. This point is alluded to indirectly via the max-crossing problem, but is not explicitly stated to the best of our knowledge.

\paragraph{Geometric intuition: }The geometric intuition from these results on convex functions with finite global minima is worth noting: there is guaranteed to exist a non-empty set $\mathcal{H}$ of supporting hyperplanes passing through the point on the epigraph corresponding to the global minimum. These hyperplanes are characterized by three fundamental properties: they contain the epigraph in their upper half-space, cannot be vertical (as this would "slice" the epigraph into two parts), and among all hyperplanes containing the epigraph in their upper half-space, achieve the highest possible intercept with the $(n+1)$st axis (as established by weak duality).

A standard result in convex analysis is that all closed proper convex functions show strong duality (see Section 4.2.1 in \cite{bertsekas2009convex} for a demonstration of this fact using the MC/MC framework). Our conditions are stronger than this condition as shown in the following proposition:

\begin{proposition}\label{prop:closed_proper}
Any closed proper convex function satisfies the conditions shown above.
\end{proposition}
\begin{proof}
Since the closedness and lower semicontinuity of $f$ are equivalent (Proposition 1.1.2 in \cite{bertsekas2009convex}), the above results follow as long as $w^*$ is finite. The properness of a closed, convex $f$ can be argued in two ways:
\begin{enumerate}
\item A closed improper convex function cannot take finite values (Page 11 of \cite{bertsekas2009convex}), and therefore, $f$ has to be proper.
\item Since $f$ has a finite global minimum, we have $f(x) > -\infty$ for all $x$, and a finite value for at least some $x$. Therefore, $f$ is proper.
\end{enumerate}
\end{proof}
To demonstrate that our result is indeed stronger, consider the function
\be
    \widehat{f(x)} = \begin{cases}
                        0 & \text{ if $x \in (-1,1)$} \\ 
                       \infty & \text{otherwise},
                  \end{cases}. \enum{clcounter}
\ee
This function above satisfies the conditions above, but is not closed. The reason our result is stronger is because lower semi-continuity is a local property of the function, whereas closedness is a global property of the function. More precisely, closedness implies lower semi-continuity at $\textit{every}$ point in the domain (Proposition 1.1.2 in \citet{bertsekas2009convex}). However as we show in Proposition \ref{prop:lsc}, our results require lower semi-continuity at only one point -- 0.

\paragraph{Connection to conjugate function theory:}The above proposition connects naturally to conjugate function theory. As shown in Section 4.2.1 of \cite{bertsekas2009convex}, for the MC/MC framework we have $w^{*} = f(0)$ and $q^{*} = f^{**}(0)$, providing another perspective on our result: for closed proper convex functions, the equality $q^* = w^*$ aligns with the fundamental property that $f(x) = f^{**}(x)$ for all $x \in \mathbb{R}^n$. By enforcing some mild constraints on the nature of $f$, we can provide an even stronger qualification of the dual optimal solution as is shown by the proposition below: 

\begin{proposition}\label{prop:compactness}
    Let $f$ be a convex function with a finite global minimum at $0$. Then the set of solutions for the max crossing problem are non empty and compact if and only if $f$ is continuous around 0.
\end{proposition}
\begin{proof}
    First suppose $f$ is continuous at 0. Non emptiness is established in the previous propositions. To establish compactness (Propositions 4.4.1 and 4.4.2 in \citet{bertsekas2009convex}) the following three conditions need to be satisfied:
    \begin{itemize}
        \item $w^{*} > -\infty$,
        \item $\overline{M} = M + \left\{(0,w) | w \geq 0 \right\}$ is convex, and
        \item The origin is an interior point of the set $D = \left\{u \text{ | there exists }w \in R \text{ with } (u,w) \in \overline{M}\right\}$.
    \end{itemize}
    
    The first condition follows immediately since \( f \) has a finite minimum at 0. The second condition holds because \( M = \text{Epi}(f) \) is convex (as \( f \) is convex), and for any \( (x, w) \in M \) and \( e \geq 0 \), we have \( (x, w + e) \in M \), so \( \overline{M} = M \). For the third condition, since \( f \) is continuous around the origin, it will be finite and continuous around the origin (since it is finite at 0). Thus, $D$ contains a ball centered at the origin, and therefore, the origin is an interior point of $D$.

    Now, for the reverse direction, suppose that the set of solutions for the max crossing problem are non-empty and compact. Then, from Proposition 4.4.2 of \citet{bertsekas2009convex}, the origin has to be an interior point of $D$ described in point number 3 above. This would require 0 to be an interior point of $dom(f) =\left\{x \in X\text{ |  }f(x) <\infty\right\}$. Since $f$ has a finite global minimum at 0, we know that 1) $f$ is proper, and 2) $0 \in dom(f)$. To establish a contradiction, suppose $f$ is discontinuous at 0. The discontinuity implies that the function has to ``jump" at 0 to a value different than $f(0)$. There are only two possibilities:
    \begin{itemize}
        \item \textbf{The function jumps to an infinite value at 0} - This is not possible because 0 would not be an interior point of $dom(f)$, which is required for compactness.
        \item \textbf{The function jumps to a finite value different than $f(0)$} - This is not possible because in this case, 0 will not be in the relative interior of $dom(f)$ (if $f$ is a proper convex function on $(-\infty,\infty]$, then $f$ restricted to $dom(f)$ is continuous over the relative interior of $dom(f)$ (Proposition 1.3.11 in \citet{bertsekas2009convex})), which would imply that 0 is not in the interior of $dom(f)$, which would imply that the solutions to the max crossing problem are not compact.
    \end{itemize}
The result follows since neither possibility is feasible.

\end{proof}
The properties established above for minimum at 0, extend naturally to any point $c \in \mathbb{R}^n$ through translation. Since translation moves points parallel to coordinate axes without rotation, it preserves all set properties, and the geometric relationships between hyperplanes and the epigraph, particularly the non-verticality property. We formalize this in the following proposition.

\begin{proposition}\label{prop:translation}
Let $f$ be a convex function with a finite global minimum at $c$. Then $q^* = w^* = f(c)$. Furthermore, if $f$ takes finite values on an open ball around $c$, then the set of optimal solutions of the max crossing problem will be non empty and compact. 
\end{proposition} 
\begin{proof}
Through the change of variable $y = x - c$, we obtain $f(x) = f(y + c) = g(y)$, which is convex in $y$ by composition. Now, $g$ is a convex function with global minimum at 0, and the results follow from our previous propositions.
\end{proof} 

The following general result follows from the above proposition.
\begin{proposition}
    Suppose a continuous, convex function, $f$, has a finite global minimum. Then the solution to the max crossing problem will be non-empty and compact.
\end{proposition}
\begin{proof}
    Special case of the previous proposition.
\end{proof}
As an application of the above proposition, we note that the set of optimal solutions to the max crossing problem for \eref{las01} will be non empty and compact.
Slater's condition follows trivially in our setup as shown in the proposition below
\begin{proposition}
    $f$ as above satisfies Slater's condition.
\end{proposition} \label{prop:slater}
\begin{proof}
    Since we are using the extended value formulation, the domain of $f$ is $\mathbb{R}^{N}$, which has two consequences:  
    (1) The affine hull of the domain is $\mathbb{R}^{N}$, meaning the relative interior coincides with the usual interior. More specifically, all feasible points with finite coordinates belong to the relative interior of the domain.
    (2) We do not have explicit inequality constraints, so any point in the relative interior of $f$ is  strictly feasible if the point is feasible.  
    Since $f(c) = w^{*} < \infty$, the point $x = c$ is both feasible and in the relative interior of $\mathbb{R}^{N}$ , thus satisfying Slater's condition.
\end{proof}

Furthermore, examining our proofs reveals that the non-verticality argument depends only on the coordinate directions in $\mathbb{R}^n$, independent of any properties of the function itself. This leads to a striking result:

\begin{proposition}\label{prop:hypzero}
For any function $h: \mathbb{R}^n \rightarrow \mathbb{R} \cup \{-\infty, \infty\}$ with a finite global minimum at $\beta = c$, there cannot exist a vertical hyperplane passing through $(c,h(c))$ containing $\text{Epi}(h)$ in its upper half-space.
\end{proposition} 
\begin{proof}
The contradiction in the non-verticality proof arises purely from the geometry of the hyperplane and the points $u$ and $-u$ constructed in $\mathbb{R}^n$. The values of $h$ at these points are irrelevant to the argument.
\end{proof}

This result requires nothing about the function beyond the existence of a finite value at the minimum. The geometry alone precludes vertical supporting hyperplanes. We can generalize the logic even further to incorporate any function whose domain is $R^{N}$ as seen in the proposition below.
\begin{proposition}\label{prop:hypgen}
    For any extended value function, $f(x)$, and any finite value $u$, there cannot exist a non-vertical hyperplane passing through $(u,f(u))$ which contains $Epi(f)$ in its positive halfspace.
\end{proposition}
\begin{proof}
    As we discussed in the geomtric proof of Proposition \ref{prop:hyperplane}, the non-existence of non-vertical supporting hyperplane at a point is only predicated on the existence of symmetric points on either side of that point in the domain. The result follows since the domain of $f$ is $R^{N}$, and $u$ is finite.
\end{proof}
\paragraph{Geometry in action:}To exemplify the generality of our geometric approach, we ``work through an example" using a variant of the Dirichlet function which is of historical importance. The function is defined as 

\be
    f_d(x) = \begin{cases}
                        0 & \text{ if x is rational} \\ 
                       1 & \text{otherwise},
                  \end{cases}. \enum{dirich1}
\ee

This is the ``worst case scenario" we could conceive of for Proposition \ref{prop:hypzero} -- a function with a global minimum at 0 with no continuity, differentiability or convexity to hang its coat on. The proposition claims that a non vertical supporting hyperplane  supporting $Epi(f_d)$ at 0 will not exist if there are points on either side of 0 belonging to the domain, a condition which is clearly being satisfied -- for any finite $u$, $f_d(u)$ and $f_{d}(-u)$ exist. This claim is easily verified for this 1-D case -- a vertical hyperplane on $Epi(f_d)$ at 0 would split the epigraph into two symmetric halves,  with the positive and negative values of $x$ lying on different sides of the hyperplane. Similarly, Proposition \ref{prop:hypgen} guarantees that the ``vanilla" Dirichlet function,
\be
    f'_d(x) = \begin{cases}
                        1 & \text{ if x is rational} \\ 
                       0 & \text{otherwise},
                  \end{cases}, \enum{dirich2}
\ee
cannot have a non-vertical hyperplane at 0 containing $Epi(f'_d)$ in its positive halfspace.

\subsubsection{Geometric properties at non-differentiable global minima}
The geometric properties discussed above take on striking implications for the loss function described in \eref{las01}.
At a differentiable minimum, the epigraph forms a smooth valley, allowing supporting hyperplanes to transition
gradually as they pivot around the minimum. This smooth structure enables gradients to follow a continuous,
predictable path down to the minimum, aligning with classical optimization techniques.
In stark contrast, when non-differentiability exists at the minimum, the epigraph assumes a distinct, angled
structure. Here, supporting hyperplanes encounter a geometric break, forcing abrupt transitions rather than a
gradual "roll" around the minimum. This creates an environment where traditional gradient-based methods
falter—they depend on the very smoothness that’s absent here.
Consider the family of supporting hyperplanes $\mathcal{H}$ at this non-differentiable minimum. While hyperplanes still
support the epigraph and are never vertical (from our general theory), their orientation shifts suddenly rather than
smoothly around the minimum point. This discontinuity reveals a fundamental barrier: without a smooth path,
gradient-based methods lack the guidance needed to navigate effectively. Thus, this geometric characterization
highlights why optimization strategies rooted in smooth settings often struggle in non-differentiable landscapes,
shedding light on the limitations encountered near such minima.

\subsubsection{A plausible explanation for the non-convergence of the ReLU networks with the Capture Theorem}

Our analysis of the dual in the previous subsection gives us some possible insights into why the NDGM sequence bounces between two bands. For completeness, we restate our core problem: we are interested in minimizing the loss function
\begin{align}
\beta^{*} 
= \arg\min_{\beta} P(\beta) 
= \arg\min_{\beta} \left\| \mathbf{y}_1 - \max(0, \mathbf{Z}_1 \beta) \right\|^2 
+ a_1 \|\beta\|_1 + a_2 \|\beta\|_2^2.
\label{eq:prel0_min}
\end{align}

where ${\bf{y}}_{1}$ is some non-positive vector, ${\bf{Z}}_{1}$ is some data matrix, $a_{1} \geq 0$, $a_{2} \geq 0$, and $a_{1} + a_{2} > 0$. This problem is convex and has a unique global optimizer of the vector of zeros. Accordingly, we could obtain the minimum by solving the following equivalent problem
\begin{equation}
\begin{aligned}
\beta^{*} 
&= \arg\min_{\beta} Q(\beta) \\
&= \arg\min_{\beta} \left\| \mathbf{y}_1 - \max(0, \mathbf{Z}_1 \beta) \right\|^2 + a_1 \|\beta\|_1 + a_2 \|\beta\|_2^2,\\
&\quad \text{subject to} \quad \|\beta\|_1 \leq 0, \quad \|\beta\|_2^2 \leq 0.
\end{aligned}
\label{eq:constrained_Q}
\end{equation}
In the extended value formulation, $Q(\beta)$ takes a finite value only at \( \beta = 0 \), and is infinite elsewhere, demonstrating the equivalence to \ref{eq:prel0_min}. The Epigraph of $Q(\beta)$ is given by
\[
E = \operatorname{epi}(Q) = 
\begin{cases}
(0, c) & \text{for all } c \geq \|\mathbf{y}\|^2, \\
(k, \infty) & \text{for all } k \neq 0,
\end{cases}
\] 
which is a vertical line along the $(n+1)^{st}$ axis. $E$ satisfies the condition of \ref{prop:bertlsc}, and therefore strong duality holds. An example of one such non-vertical hyperplane is the affine space parallel to first $n$ axes, which passes through ($0,0...0, ||y|^{2}$). The Lagrangian for $Q(\beta)$  will be given by 
\begin{align}
R(\beta) 
=  \left\| \mathbf{y}_1 - \max(0, \mathbf{Z}_1 \beta) \right\|^2 
+ a'_1 \|\beta\|_1 + a'_2 \|\beta\|_2^2,
\label{eq:lagr}
\end{align}
where $a_{1}'$ and $a_{2}'$ incorporate the Lagrange multipliers in $a_{1}$ and $a_{2}$. Note that this is algebraically identical to the original unconstrained objective in Equation~\ref{eq:prel0_min}, highlighting that our loss can be interpreted as the Lagrangian associated with a constraint set containing only the origin. The dual function corresponding to $R(\beta)$ is
\be
d(a'_{1} - a_{1}, a'_{2} - a_2) = ||\mathbf{y}_1 ||^{2},
\ee
which is independent of the Lagrange multiplier. There are two conclusions from this: first, the set of dual solutions is unbounded, and is therefore not compact corroborating Proposition \ref{prop:translation} since our modified loss function is not continuous at 0, and second that there is no dual "force" anchoring the iterates towards the optimal solution. This also corroborates with Proposition \ref{prop:compactness}; since We hypothesize that this is the reason why our NDGM sequence in our Capture Theorem plots are bouncing between "equally plausible" points. To use the "taxation" analogy of the dual variables (section 5.4.4 in \cite{boyd2004convex}), the price of violation becomes irrelevant when nothing is stored in the warehouse!

\section{An Axiomatic Anthology of Numerical Alliterations}

\subsection{Proof of proposition \ref{prop:lassooverall}}
\label{app:lassooverall}

Proposition \ref{prop:lassooverall} follows from the two propositions described below.

\begin{proposition} \label{prop:lasso1}
Let $\{\beta_{t}\}_{t=0}^\infty$ denote the sequence generated by running NDGM on \eref{las01}, with a constant learning rate, $\alpha$, and the entries in $\beta_{0}$ being uniformly sampled from $[-1,1]$. Then $\beta_{k}$ will have all non-zero entries with probability 1 for all $k > 0$.
\end{proposition}
\begin{proof}
The NDGM iteration for \eref{las01} is described by the recursion
\be 
\beta_{i}[k] =   \begin{cases}
                        \beta_{i-1}[k] - \alpha\lambda_{1} & \text{ if $\beta_{i-1}[k] > 0$}\\
                       \beta_{i-1}[k] + \alpha\lambda_{1} & \text{ if $\beta_{i-1}[k] < 0$}\\
                        0 & \text{ if $\beta_{i-1}[k] = 0$} \enum{gd1}
                  \end{cases}.
\ee 
From \eref{gd1}, it is clear that if the $k^{th}$ entry in the vector becomes zero during any iteration, then it stays zero for all subsequent iterations. Accordingly, suppose the $k^{th}$ entry in the NDGM sequence vector becomes 0 for the first time after $N$ iterations. Then we have
\be
\beta_{N}[k] = \beta_{0}[k] + (N-2m)\alpha\lambda_{1} = 0, \enum{las1}
\ee 
where $m$ is the number of times the $k^{th}$ entry of the parameter exceeds 0 in the first $N$ iterations. \eref{las1} can only hold if $\beta_{0}[k]$, is an integer multiple of $\alpha\lambda_{1}$. Since the feasible values are countable, and the set of initializations is uncountable, the probability of the occurrence has measure 0.
\end{proof}
In the next proposition, we show that the sequence does not converge, but oscillates between two fixed points.

\begin{proposition}\label{thm:proplasso2}
    The sequence $\{\beta_{t}\}_{t=0}^\infty$ described in the previous proposition does not converge. Furthermore, there exists an integer $N_{0}$, such that for every $N \geq N_{0}$, we can write

    \be
    \beta_{N + m}[k] = \begin{cases}
                        \gamma_{k} - \alpha\lambda_{1} & \text{ if m is odd} \\ 
                       \gamma_{k} & \text{ if m is even},
                  \end{cases} \enum{lassooverallrepeat}
\ee
for some $\gamma_{k} \in (0,\alpha\lambda_{1})$ for all $0\leq k\leq P-1$.
\end{proposition}
\begin{proof}
We prove the results assuming $\beta_{0}[k] > 0$; the proof is similar when $\beta_{0}[k] < 0$. From Proposition \ref{prop:lasso1}, we know that for any $N > 0$, $\beta_{N}[k] \neq 0$ with probability 1. Therefore, \eref{gd1} implies that  starting from any $\beta_{0}[k] > 0$, the  sequence will decrease monotonically till it reaches a value between $0$ and $\alpha\lambda_{1}$. Accordingly, let $\beta_{n_{k}}[k] = \gamma_{k}$ for some $n_{k} \geq 0$, with $\gamma_{k} \in (0,\alpha\lambda_{1})$. Since $\gamma_{k} - \alpha\lambda_{1} < 0$, completing the recursion in \eref{gd1} gives
\be
    \beta_{n_k + m}[k] = \begin{cases}
                        \gamma_{k} - \alpha\lambda_{1} & \text{ if m is odd} \\ 
                       \gamma_{k} & \text{ if m is even}.
                  \end{cases} \enum{lasso1}
\ee

The result follows by choosing $N_{0} = \text{max}(n_{0}, n_{1} \hdots n_{P-1})$.
\end{proof}

\subsection{NDGM unreliability for the general LASSO problem} \label{app:lassogeneral}

In this section, we demonstrate via simulations that the show that the problems demonstrated in our toy LASSO problem hold even for the general setting. Consider the loss function for the general LASSO problem given by
\be 
L_{4}(\beta;{\bf{W}},{\bf{y}},\lambda_{1}) = \frac{1}{N}||{\bf{y}} -{\bf{W}}\beta||_{2}^{2} + \lambda_{1}||\beta||_{1}, \enum{laspar1}
\ee
where ${\bf{W}}$ is an arbitrary $20 \times 500$ dimensional data matrix, and ${\bf{y}}$ is an arbitrary $20 \times 1$ response, each of whose entries are sampled from the uniform distribution on $[-1, 1]$. We run NDGM on \eref{laspar1} twice with the same learning rate and initialization, once with $\lambda_{1} = 0.01$, and once with $\lambda_{1} = 10$. At the end of the two runs, the optimal solution with $\lambda_{1} = 0.1$ has an $L_{1}$ norm of 1.62, and the optimal solution with $\lambda_{1} = 10$ has an $L_{1}$ norm of 25.4 i.e., a 1000 fold increase in the value of $\lambda_{1}$ results in a more than 15 fold increase in the $L_{1}$ norm of the optimal solution!

\subsection{Different NGDM variants and the LASSO} 
\label{app:ndgmvariants}

In this section, we demonstrate that various NDGM variants, traditionally thought to converge similarly based on the analysis of differentiable functions, exhibit notably distinct behaviors even when applied to the simple non-differentiable toy LASSO problem outlined in \eref{las01}. A summary of our findings are in Table \ref{tab:variantcomp}.

\begin{table}[ht]
    \centering
    \begin{tabular}{|m{4cm}|m{5cm}|}
        \hline
        \textbf{NDGM variant} & \textbf{Unique characteristic}  \\ \hline
        Vanilla NDGM (including SGD) &  $\beta_{0}$ and $\lambda_{1}$ influence eventual values of the sequence. If sequence hits 0, then stays 0.\\ \hline
        NDGM with momentum & If sequence hits 0, then jumps away from 0.\\ \hline
        RMSProp & Only $\beta_{0}$ influences the eventual value of the sequence. If sequence hits 0, then stays 0.\\ \hline
        \hline
        Adam &  If sequence hits 0, then bounces away from 0. RMSProp like behavior in certain limits\\ \hline
    \end{tabular}
    \caption{Unique characteristics of sequences generated on running the different NDGM variants on the LASSO problem described in \eref{las01} with an initialization of $\beta_{0}$, and a LASSO penalty of $\lambda_{1}$. These sequences are believed to have similar behavior based on the analysis of differentiable loss functions.}
    \label{tab:variantcomp}
\end{table}

\subsubsection{SGD and the LASSO}
Since the gradient of \eref{las01} is independent of the batch in question, the results in Proposition \ref{thm:proplasso2} hold for the SGD case as well.

\subsubsection{NDGM with momentum and the LASSO}

With the vanilla NDGM, \eref{gd1} gives the guarantee that once a parameter value reaches zero during an iteration, it remains zero in all future iterations. We show that NDGM with momentum does not provide this guarantee.

\begin{proposition} \label{prop:lassomom}
Let $\{\beta_{t}\}_{t=0}^\infty$ denote the sequence generated by running NDGM with momentum on \eref{las01}, with a constant learning rate, $\alpha$, and the entries in $\beta_{0}$ being uniformly sampled from $[-1,1]$. Suppose $\beta_{N-1}[k] \neq 0$ and $\beta_{N}[k] = 0$, then $\beta_{N+1}[k] \neq 0$.
\end{proposition}

\begin{proof}
    With a momentum factor of $\eta$, NGDM with momentum can be written as (see page 6 of \url{https://www.ceremade.dauphine.fr/~waldspurger/tds/22_23_s1/advanced_gradient_descent.pdf} for details) 
    \be
    \beta_{N+1}[k] = \beta_{N}[k] - \alpha(1- \eta) \nabla L_{3}(\beta_{N})[k] + \eta(\beta_{N}[k] - \beta_{N-1}[k]) \enum{gdwm1}
    \ee
Since $\beta_{N}[k] = 0$, we have $\nabla L_{3}(\beta_{N}[k]) = 0$, and therefore, 
\be 
\beta_{N+1}[k] = -\eta \beta_{N-1}[k] \neq 0, \enum{gdwm2} 
\ee 
hence the result.    
\end{proof}

\subsubsection{RMSProp and the LASSO}
Proposition \ref{thm:proplasso2} shows that the vanilla NDGM sequence eventually bounces between two vectors which depend on the strength of regularization, $\lambda_{1}$, and the weight initialization, $\beta_{0}$. Here we show that RMSProp, which is a near scale invariant NDGM  sequence bounces between two vectors whose orientation depends only on $\beta_{0}$. The RMSProp equations (equation 18 in \citep{ruder2016overview}) can be written as
\begin{align}
v_{t}[k] &= \gamma v_{t-1}[k] + (1-\gamma)\left(\nabla L_{3}(\beta_{N})[k]\right)^{2}, \mbox{ and }  \enum{eq:rms1} \\ 
\beta_{t+1}[k] &= \beta_{t}[k] - \frac{\alpha}{\sqrt{v_{t}[k] + \bar{\epsilon} }}\nabla L_{3}(\beta_{t})[k] \enum{eq:rms2}, 
\end{align}
where $v_{0}[k] = 0$, $\gamma$ is a scaling factor with a typical default value of 0.99, and $\bar{\epsilon}$ is a small positive jitter added to prevent division by 0. If $\beta_{N}[k] \neq 0$, we have  $\left(\nabla L_{3}(\beta_{N})[k]\right)^{2} = \lambda_{1}^{2}$. Using this fact in \eref{eq:rms1}, we note that $v_{1}[k] = (1-\gamma)\lambda_{1}^{2}$, and  $v_{2}[k] = (1-\gamma^{2})\lambda_{1}^{2}$. Completing the recursion we get
\be 
v_{t}[k] = (1 - \gamma^{t})\lambda_{1}^{2} \enum{eq:rms301}.
\ee 
Plugging \eref{eq:rms301} into \eref{eq:rms2} with $\epsilon = \bar{\epsilon}/\lambda_{1}^{2}$ gives
\be 
\beta_{t+1}[k] = \beta_{t}[k]- \frac{\alpha}{\sqrt{1 + \epsilon - \gamma^{t}}} sign(\beta_{t}[k]), \enum{rmsprop}
 \ee
which we use as a base in our analysis. We now provide the key results for RMSProp:
\begin{proposition}
Let $\{\beta_{t}\}_{t=0}^\infty$ denote the sequence generated by running the RMSProp recursion described in \eref{rmsprop}, with a constant learning rate, $\alpha$, and the entries in $\beta_{0}$ being uniformly sampled from $[-1,1]$. Then $\beta_{t}$ will have all non-zero entries with probability 1 for all $t > 0$.
\end{proposition} \label{prop:rp1}
Suppose the the $k^{th}$ entry in the RMSProp sequence vector becomes 0 for the first time after $N$ iterations. Then, completing the recursion in \eref{rmsprop} we have 
\be 
\beta_{N}[k] = \beta_{0}[k] + \sum_{i \in  Q} \frac{\alpha}{\sqrt{1 + \epsilon - \gamma^{i}}} -  \sum_{j \in P} \frac{\alpha}{\sqrt{1 + \epsilon - \gamma^{j}}} \enum{eq:rms3}
= 0,
\ee 
where $Q$ is the set of all iteration numbers where the parameter value is less than 0 and $P$ is the set of all iteration numbers where the parameter value is greater than 0, with $|P|+ |Q| = N$. Let $C(N)$ denote the set of all values of $\beta_{0}[k]$ which satisfy \eref{eq:rms3}. The cardinality of $C(N)$ is atmost equal to the number of ways of partitioning $N$ integers into 2 sets, $P$ and $Q$. Therefore, $C(N)$ is countable, and the superset of all feasible values of $\beta_{0}[k]$ which lead to a sparse solution given by $\bigcup_{I = 1}^{\infty} C(I)$ is countable. Since the initializations are uncountable, the probability of the event has measure 0.

 \begin{proposition}
     The RMSProp sequence described in \eref{rmsprop} eventually behaves like a vanilla NGDM sequence with a LASSO penalty of 1 irrespective of the value of $\lambda_{1}$ chosen in the original problem.
 \end{proposition} \label{prop:rmsfinal}
 \begin{proof}
From the previous proposition, we know that $\beta_{t}[k] \neq 0$ with probability 1 even for large values of $t$. Furthermore, we note that even after a modest number of iterations the scaling term becomes negligible; for instance, with $t = 1000$, $\gamma^{t} = O(10^{-5})$. Accordingly, assuming that in the large $t$ limit we have $ \epsilon \ll \gamma^{t} \ll 1$, we can use first order Taylor expansions to write
\be 
\beta_{t+1}[k] = \beta_{t}[k]- \alpha sign(\beta_{t}[k]) + O(\gamma^{t}), \enum{rmspropfinal}
\ee
which is the recursion for the vanilla NGDM algorithm with a LASSO penalty of 1 up to first order.
 \end{proof}

 \subsubsection{Adam and the LASSO}
The Adam recursion is given by
\begin{align}
v_{t}[k] &= \gamma v_{t-1}[k] + (1-\gamma)\left(\nabla L_{3}(\beta_{t})[k]\right)^{2},   \label{eq:adam1} \\ 
m_{t}[k] &= \delta m_{t-1}[k] + (1 - \delta)\nabla L_{3}(\beta_{t})[k], \label{eq:adam2}\\ 
\beta_{t+1}[k] &= \beta_{t}[k] - \frac{\alpha}{\sqrt{v_{t}[k] + \bar{\epsilon} }}m_{t}[k], \label{eq:adam3}
\end{align}
where the scaling terms are as described for RMSProp, and $m_{t}[k]$ is a momentum term with $m_{0}[k] = 0$. We now derive the key properties of the Adam recursion.

\begin{proposition}
    $|m_{t}[k]| < \lambda_{1}$ for all $t$ and $k$. 
\end{proposition} \label{adam:prop1}
\begin{proof}
    From \ref{eq:adam2}, we observe that $m_{t+1}[k]$ is a convex combination of $m_{t}[k]$ and either $-\lambda_{1}$ or $\lambda_{1}$. By induction, if $|m_{t}[k]| < \lambda_{1}$, then $|m_{t+1}[k]| < \lambda_{1}$. Since $m_{0}[k] = 0$, the result follows.
\end{proof}

\begin{proposition}
Suppose $\delta$ is uniformly sampled from [0.9, 0.99]. Then $m_{t}[k] \neq 0$ with probability 1 for all $t>0$.   
\end{proposition}
\begin{proof}
    Suppose $m_{T}[k] = $. Using $m_{0}[k] = 0$, we can write \ref{eq:adam2} as 
    \be 
     m_{T}[k] = (1-\delta)\sum_{i=1}^{i=T} \operatorname{sign}(\beta_{i}[k])\delta^{i-1}. \enum{eq:adammom}
    \ee 
    The polynomial equation $m_{T} = 0$ has at most $T$ roots, and is therefore countable. Since the sample space is uncountable, the probability of the event has measure 0.
\end{proof}

An interesting corollary of the above result is that for $\delta$ as above, $\beta_{T}$ will not be sparse.

\begin{proposition}
    Suppose $\delta$ is some fixed value between [0.9, 0.99], and $\beta_{0}$ is uniformly sampled from [-1, 1]. Then $\beta_{T}$ will be non-sparse with probability 1 for all $T > 0$.
\end{proposition}
\begin{proof}
For our proof we use $\alpha_{t} = \frac{\alpha}{\sqrt{v_{t}[k] + \bar{\epsilon} }}$, noting that $\alpha_{t} > 0$ for all $t > 0$, and is independent of $\beta_{0}$ and $\delta$. Suppose the $k^{th}$ entry in the vector becomes 0 after $N+1$ steps. Then, completing the recursion in \ref{eq:adam3} using \eref{eq:adammom} with $m_{0}[k] = 0$ gives
\begin{align}
    0 = \beta_{N+1}[k] &= \beta_{0}[k] - \sum_{i=0}^{N}\alpha_{i}m_{i}[k] \\ 
    &= \beta_{0}[k] -(1-\delta) \sum_{i=1}^{i=N}\sum_{j= 1}^{j=i} \alpha_{i} \operatorname{sign}(\beta_{j}[k])\delta^{j-1}. \enum{eq:adam:nonsparse}
\end{align}
Since $\operatorname{sign(\beta_{j}[k]})$ can take only one of two values for each $j$, the above polynomial equation can have atmost $2^{N}$ permutations for its coefficients. Accordingly, the set of $\beta_{0}$ values which can lead to sparsity has cardinality at most $N \times 2^{N}$, and is therefore countable. Since the initializations are uncountable, the probability of the event has measure 0.
\end{proof}

The next proposition shows that in the unlikely event the coefficient hits 0 on an iteration, it will bounce away from it in the next iteration.

\begin{proposition}
    Let $\{\beta_{t}\}_{t=0}^\infty$ denote the sequence generated by running Adam on \eref{las01}, with a constant learning rate, $\alpha$, and the entries in $\beta_{0}$ being uniformly sampled from $[-1,1]$. If $\beta_{N}[k] \neq 0$, then $\beta_{N+1}[k] \neq 0$.
\end{proposition}
\begin{proof}
    Since $\beta_{N}[k] = 0$, we have $\nabla L_{3}(\beta_{N}[k]) = 0$, and therefore, from \ref{eq:adam2}, $m_{N+1}[k] = \delta m_{N}[k]$. The result follows by substituting $m_{N+1}[k]$ into \ref{eq:adam3}.
\end{proof}

Since $\beta_{0}[k] > 0$, we have $m_{0}[k] > 0$. Furthermore, if the $k^{th}$ entry in the vector remains positive for the first $T$ steps, then $m_{T}[k] > 0$. This is generalized in the next proposition. Define a "run" of size $T$ for $\beta_{j}[k]$ if $\operatorname{sign}(\beta_{j}[k]) = \operatorname{sign}(\beta_{i+j}[k])$ for all $i < T$. That is, for the duration of a run, the sign remains unchanged.

\begin{proposition}
    For a run of length $T$ starting at  $\beta_{j}[k]$, we can write
    \be 
    m_{j+T}[k] = \delta^{T}m_{j}[k] + (1 - \delta^{T})\operatorname{sign}(\beta_{j}[k])\lambda_{1}.
    \ee
\end{proposition}
\begin{proof}
    Using \ref{eq:adam2} and the definition of a run, we have
    \begin{align}
        m_{j+2}[k] &= \delta m_{j+1}[k] + (1 - \delta) \operatorname{sign}(\beta_{j}[k])\lambda_{1}, \\
                   &= \delta^{2} m_{j}[k] + (1-\delta) \operatorname{sign}(\beta_{j}[k])\lambda_{1} (1 + \delta).
    \end{align}
    Completing the recursion to $T$ yields
    \begin{align}
        m_{j+T}[k] &= \delta^{T}m_{j}[k] + (1-\delta)\operatorname{sign}(\beta_{j}[k])\lambda_{1}(1 + \delta + \dots + \delta^{T-1}) \\
        &= \delta^{T}m_{j}[k] + (1-\delta^{T})\operatorname{sign}(\beta_{j}[k])\lambda_{1}.
    \end{align}
\end{proof}

Since $\delta < 1$, this implies that for long runs, $m_{t}[k] \approx \operatorname{sign}(\beta_{j}[k])\lambda_{1}$. Using \eref{eq:rms301}, we obtain

\be 
\beta_{m+1}[k] \approx \beta_{m}[k] - \alpha \operatorname{sign}(\beta_{m}[k]),
\ee 

which mirrors the RMSprop recursion. Furthermore, from Proposition \ref{adam:prop1}, we have $|m_{t}[k]| < \lambda_{1}$. Thus, when $\beta_{m}[k]$ changes sign, its magnitude stabilizes near $\lambda_{1}$, but with an alternating direction.

\subsection{Connections to deep learning training theory: }\label{app:dltheory}
The conventional wisdom in deep learning is that different NDGM variants converge at varying rates, but eventually reach similar near-optimal values by the end of training. In this section, we demonstrate problems with this assumption using our toy LASSO problem from Equation \eref{las01}.

Since the vector of zeros is the unique global minimizer of Equation \eref{las01}, we expect the vanilla NDGM and RMSProp sequences to converge to this value by the end of training. Accordingly, we anticipate the $L_{1}$ norms of the sequences generated using these two variants to converge to comparable near-zero values after training completes. Because the loss function in Equation \eref{las01} is the $L_{1}$ norm of the weights scaled by a factor of $\lambda_{1}$, we expect the NDGM and RMSProp training curves to also hover around 0 at the end of training, exhibiting substantial overlap. To test this hypothesis, we ran vanilla NDGM and RMSProp on Equation \eref{las01} with identical initializations and learning rates, plotting the final 100 training iterations for each sequence using two different $\lambda_{1}$ values. As Figure \ref{fig:rmsvgdcomp} shows, the two sequences display no overlap for either $\lambda_{1}$ value. Moreover, the RMSProp solution consistently has a higher loss (and $L_{1}$ norm) across all 100 iterations when $\lambda_{1}$ = 0.001, while vanilla NDGM always yields a higher value when $\lambda_{1}$ = 1. Conventional analysis of differentiable functions cannot explain this peculiarity, but our findings provide intuition into the underlying reasons. To explain these findings, we use the following proposition:
\begin{figure*}[htbp]
\centering
\begin{minipage}{0.47\textwidth}
\includegraphics[width=\linewidth]{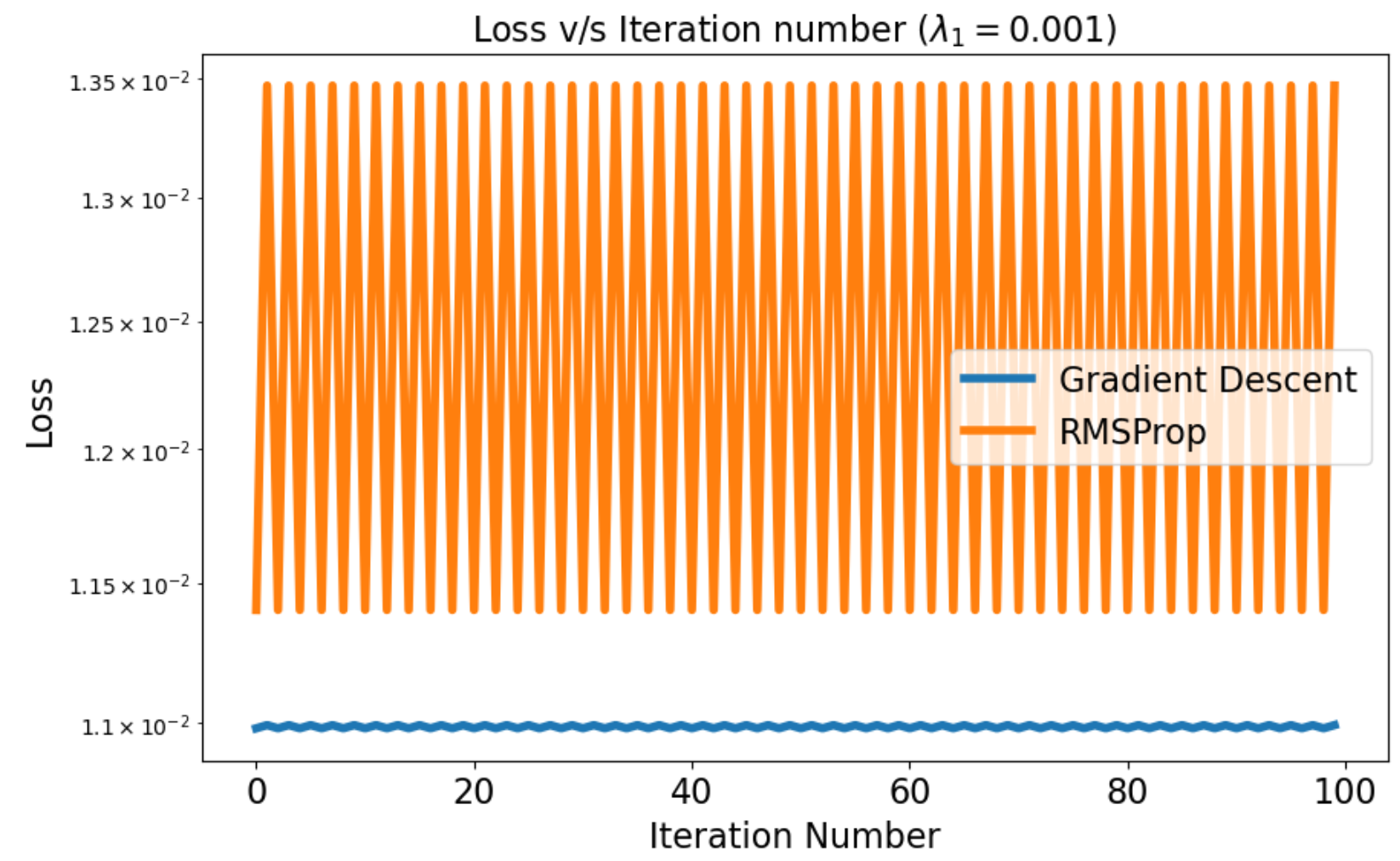}
\label{fig:rmsl001}
\end{minipage}
\hfill
\begin{minipage}{0.47\textwidth}
\includegraphics[width=\linewidth]{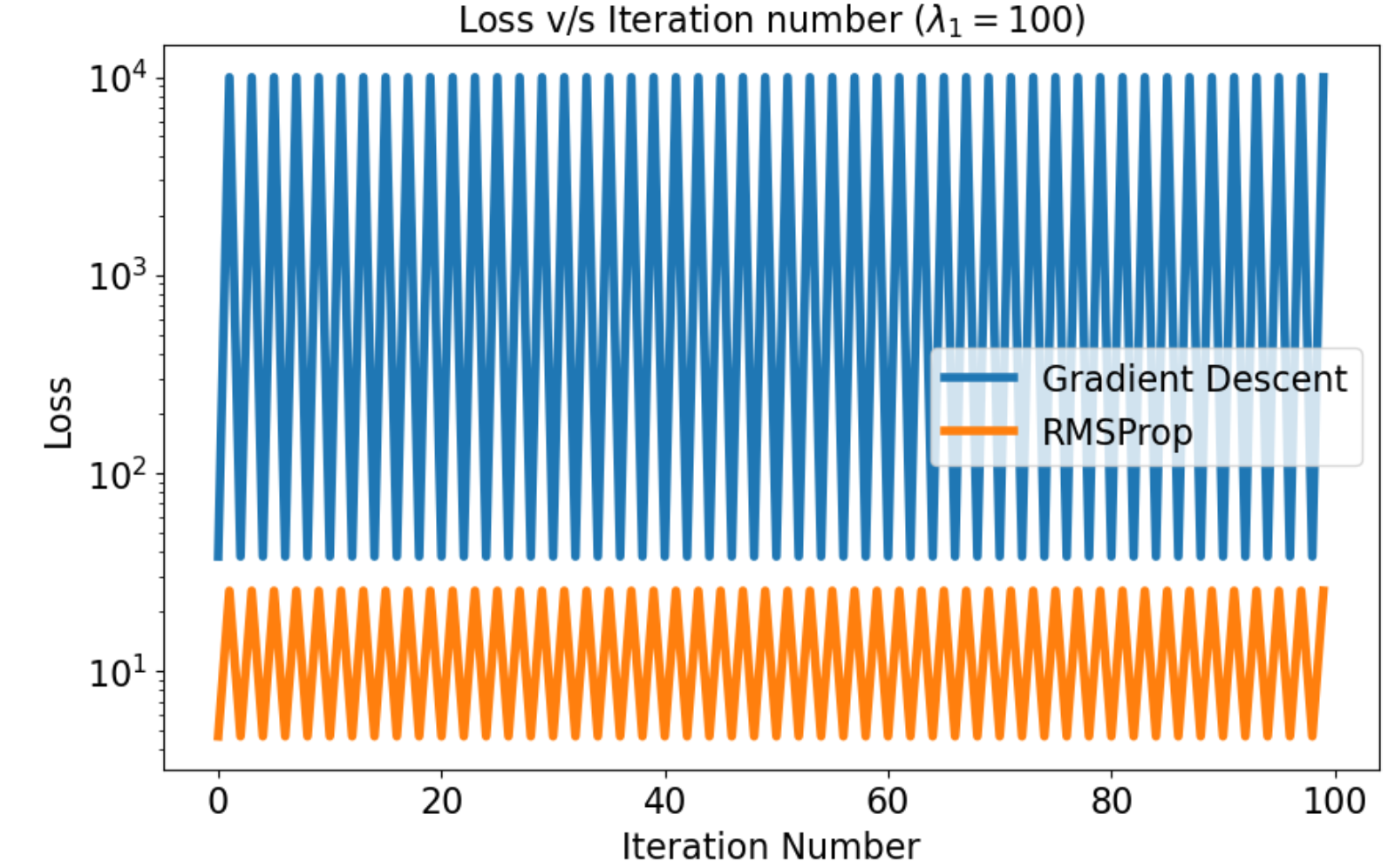}
\label{fig:rmsl100}
\end{minipage}
\caption{Comparison of the training curves of RMSProp and vanilla NDGM with $\lambda_{1} = 0.001$, and $\lambda_{1} = 100$ for the toy LASSO problem considered in \eref{las01} with $P = 10$ and $\alpha = 0.1$. We note that when $\lambda_{1} \ll 1$, the RMSProp training curve lies above the vanilla NDGM training curve, and the reverse is true for $\lambda_{1} \gg 1$.}
\label{fig:rmsvgdcomp}
\end{figure*}

\begin{proposition}
    For the vanilla NDGM sequence we have $0 < \liminf_{N \rightarrow \infty} ||\beta_{N}||_{1} \leq \limsup_{N \rightarrow \infty} ||\beta_{N}||_{1} < P\alpha\lambda_{1}$, where $P$ is the number of rows in $\beta_{0}.$
\end{proposition} \label{prop:l1norm}
\begin{proof}
    From Proposition \ref{thm:proplasso2}, we note that $\liminf_{N \rightarrow \infty}|\beta_{N}[k]| > 0$ and $\limsup_{N \rightarrow \infty}|\beta_{N}[k]| < \alpha\lambda_{1}$ for all $k$. Therefore, $0 < \sum_{i=0}^{P-1} \liminf_{N \rightarrow \infty}|\beta_{N}[i]| \leq \liminf_{N \rightarrow \infty}||\beta_{N}||_{1} \leq \limsup_{N \rightarrow \infty}||\beta_{N}||_{1} \leq \sum_{i=0}^{P-1} \limsup_{N \rightarrow \infty}|\beta_{N}[i]| < P\alpha\lambda_{1}$. 
\end{proof}

We know from propositions \ref{thm:proplasso2} and \ref{prop:rmsfinal} that the vanilla NDGM sequence and the RMSProp sequence eventually bounce (approximately) between two points each. Let $v_{1}(\beta_{0}, \lambda_{1})$ and $v_{2}(\beta_{0}, \lambda_{1})$ indicate the points around which the vanilla NDGM sequence bounces, and let $r_{1}(\beta_{0}, \lambda_{1})$ and $r_{2}(\beta_{0}, \lambda_{1})$ indicate the points around which the RMSProp sequence bounces. From propositions \ref{prop:l1norm} and \ref{thm:proplasso2} we have that $\mbox{min}(\lambda_{1}||v_{1}(\beta_{0}, \lambda_{1}||_{1}, 
 \lambda_{1}||v_{2}(\beta_{0}, \lambda_{1})||_{1}) < P\alpha\lambda_{1}^{2}$, and $\mbox{max}(\lambda_{1}||v_{1}(\beta_{0}, \lambda_{1}||_{1}), \lambda_{1}||v_{2}(\beta_{0}, \lambda_{1})||_{1}) < P\alpha\lambda_{1}^{2}$. From proposition \ref{prop:rmsfinal} we have that the RMSProp sequence  eventually behaves like a vanilla NDGM sequence with a LASSO penalty of 1, and therefore, we have $\mbox{min}(\lambda_{1}||r_{1}(\beta_{0},\lambda_{1}||_{1}, \lambda_{1}||r_{2}(\beta_{0}, \lambda_{1})||_{1}) < P\alpha\lambda_{1}$, and $\mbox{max}(\lambda_{1}||r_{1}(\beta_{0}, \lambda_{1}||_{1}, \lambda_{1}||r_{2}(\beta_{0}, \lambda_{1})||_{1}) < P\alpha\lambda_{1}$. Therefore, when $\lambda_{1} \ll 1$, we intuitively expect $\mbox{max}(\lambda_{1}||v_{1}(\beta_{0}, \lambda_{1}||_{1}, \lambda_{1}||v_{2}(\beta_{0}, \lambda_{1})||_{1}) < \mbox{min}(\lambda_{1}||r_{1}(\beta_{0}, \lambda_{1}||_{1}, \lambda_{1}||r_{2}(\beta_{0}, \lambda_{1})||_{1})$, and  when $ \lambda_{1} \gg 1$ we expect $\mbox{max}(\lambda_{1}||r_{1}(\beta_{0}, \lambda_{1}||_{1}, \lambda_{1}||r_{2}(\beta_{0}, \lambda_{1})||_{1}) < \mbox{min}(\lambda_{1}||v_{1}(\beta_{0}, \lambda_{1}||_{1}, \lambda_{1}||v_{2}(\beta_{0}, \lambda_{1})||_{1})$, which is what the figure shows. 

To clarify the ideas, suppose $P = 10$ and $\alpha = 0.1$. The above propositions state that with $\lambda_{1} = 0.01$, the RMSProp loss oscillates between two positive values, each less than 0.01, while the vanilla NDGM loss oscillates between two positive values, each less than 0.0001. Consequently, when $\lambda_{1}$ is significantly less than 1, it is reasonable to anticipate the RMSProp loss will typically be greater than the vanilla NDGM loss at training's end. However, when $\lambda_{1} = 100$, the RMSProp loss oscillates between two positive values under 100, whereas the vanilla NDGM loss oscillates between two values under 10,000. Accordingly, when $\lambda_{1}$ is significantly greater than 1, the RMSProp loss is expected to be less than the vanilla NDGM loss in most scenarios at training's conclusion, as demonstrated in the figure.

\section{Experiments with the LASSO penalty:}\label{app:lassoexp} 

In this section, we provide a table describing the percentage of parameters having absolute values less than $10^{-5}$ as a function of the algorithm used, and the strength of regularization. We also show a plot of the $L_1$ norms using a different seed (or a different weight initialization) which shows that the $L_1$ norm of the final solution is sensitive to the weight initialization making interpretability of the solution a moot point. The code is GPT generated and validated for correctness

\begin{figure}[!htbp]
\centering
\begin{minipage}{0.9\textwidth}
  \centering
  \includegraphics[width=0.48\textwidth]{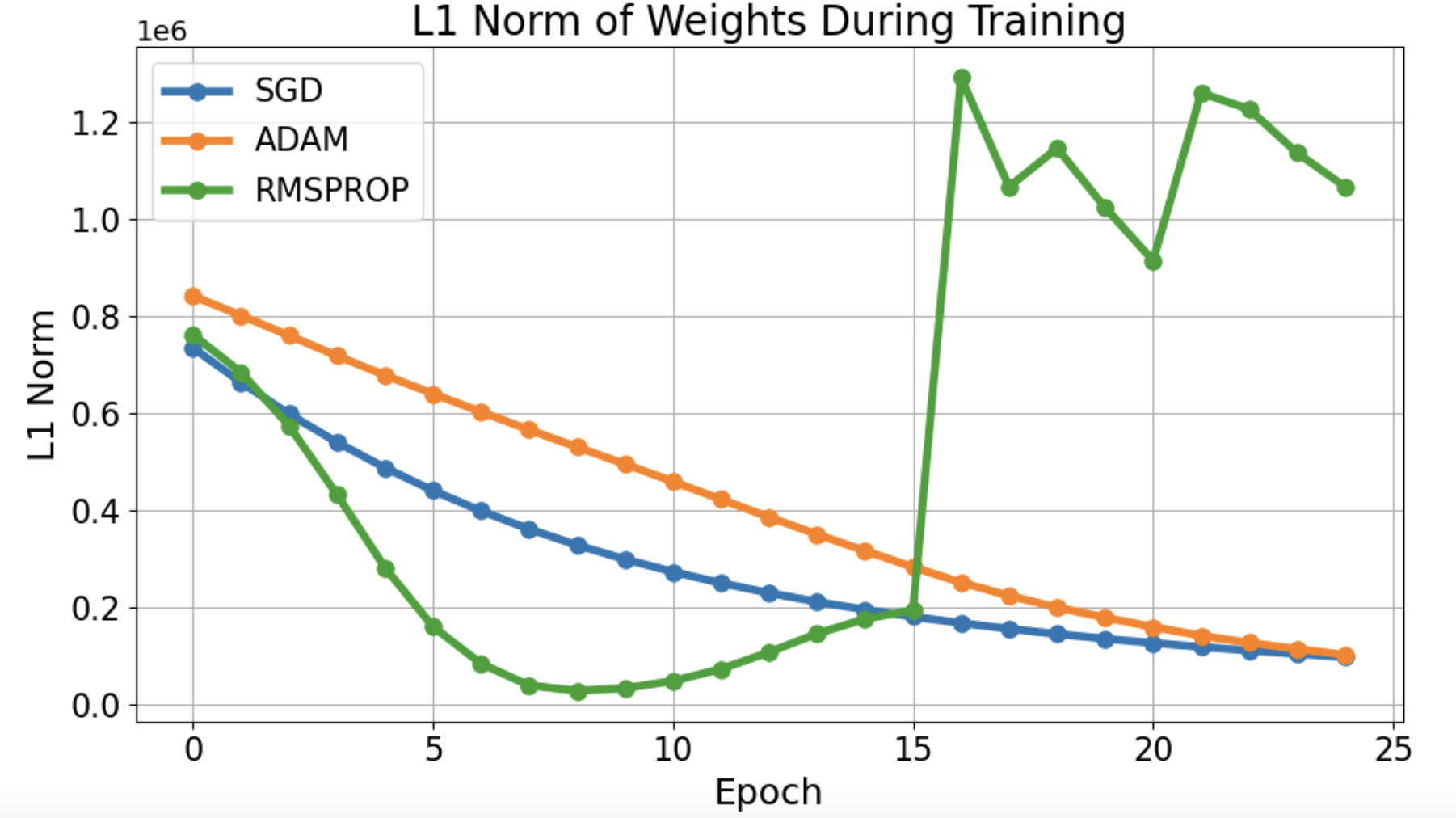}\hfill
  \includegraphics[width=0.48\textwidth]{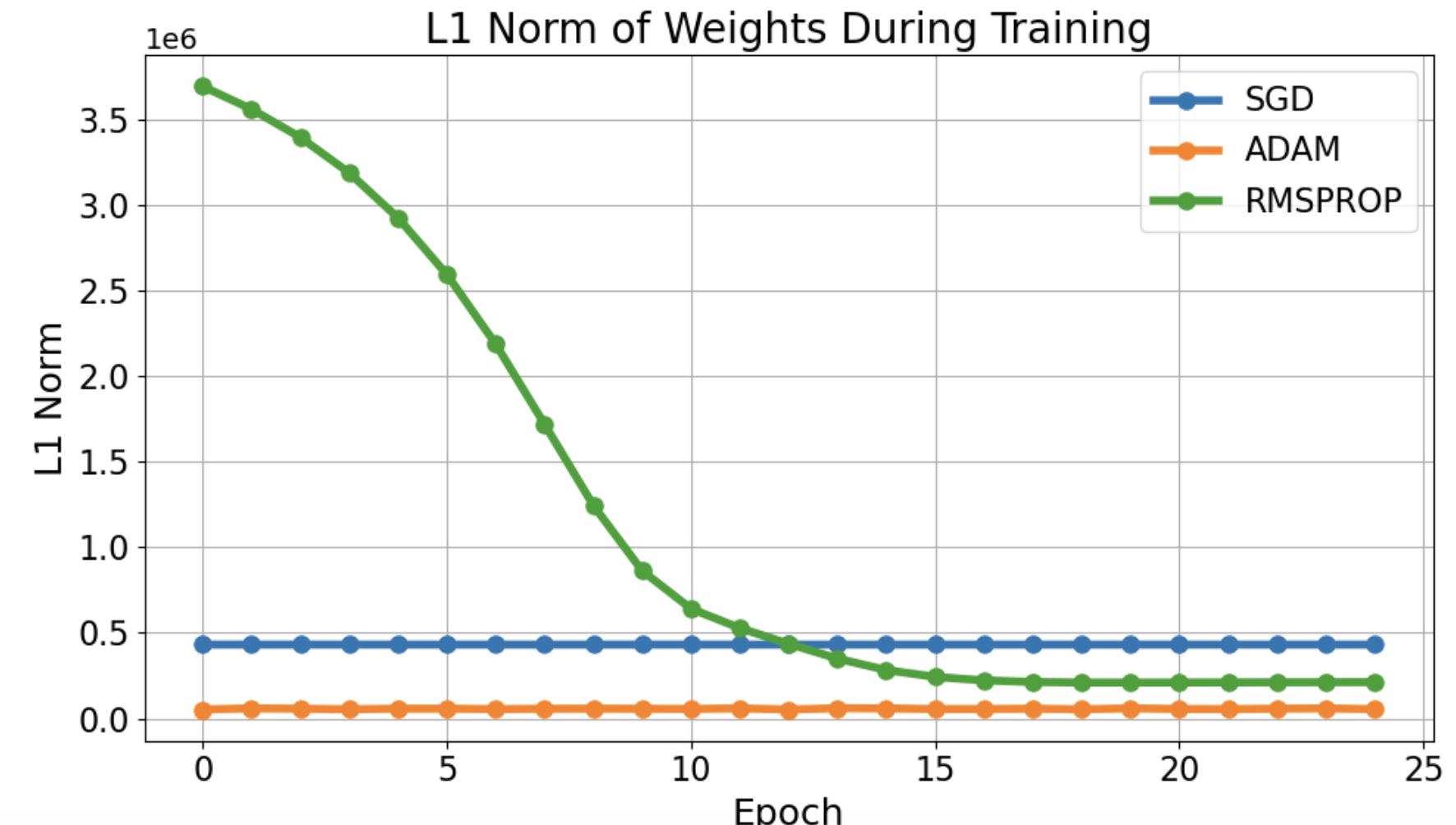}
  \vspace{0.5em}
\end{minipage}
\caption{$L_1$ norm of the weights as a function of iteration number for $\lambda_{1} = 0.002$ (left) and $\lambda_{1} = 2$ (right) for a different weight initialization. Note that the plots look very different on changing the initialization. In each figure, the green, orange, and blue lines represent curves for RMSProp, Adam, and vanilla SGD respectively.}
\label{fig:l1comparisonalt}
\end{figure}

\begin{table}[!htb]
    \centering
    \begin{tabular}{|m{3cm}|m{2cm}|m{2cm}|m{2cm}|m{2cm}|}
        \hline
        \textbf{Variant} & $\lambda_{1} = 0.002$ & $\lambda_{1} = 0.002$ (alt) & $\lambda_{1} = 2$ & $\lambda_{1} = 2$ (alt) \\ \hline
        SGD & 90.13\% & 91.34\% & 0.18\% & 0.18\% \\ \hline
        RMSProp & 0.02\% & 1.00\% & 2.61\% & 2.67\% \\ \hline
        Adam & 90.76\% & 90.68\% & 2.89\% & 2.88\% \\ \hline
    \end{tabular}
    \caption{Comparison of the percentage of weights having absolute value less than $10^{-5}$ across the three NDGM variants, for two different random initializations ("alt" denotes alternate seed runs). Sparsity outcomes are sensitive to initialization, emphasizing the instability of $L_{1}$ regularization under NDGM training. Notably, a $0.1\%$ difference in sparsity corresponds to $10,000$ variables in a model with a modest $100$ million parameters.}
    \label{tab:l1perccomp}
\end{table}

\begin{verbatim}
import torch
import torch.nn as nn
import torch.optim as optim
from torchvision import datasets, transforms, models
import copy 


# Set random seed for reproducibility
seed = 42
torch.manual_seed(seed)
torch.backends.cudnn.deterministic = True
torch.backends.cudnn.benchmark = False
import numpy as np
np.random.seed(seed)

# Set device to GPU if available
device = torch.device("cuda" if torch.cuda.is_available() else "cpu")

# Define transform and download CIFAR-10 dataset
transform = transforms.Compose([
    transforms.ToTensor(),
    transforms.Normalize((0.5, 0.5, 0.5), (0.5, 0.5, 0.5)),
])

train_dataset = datasets.CIFAR10(root='./data', 
                                 train=True, 
                                 download=True, 
                                 transform=transform)
train_loader = torch.utils.data.DataLoader(train_dataset,
                                           batch_size=512, 
                                           shuffle=False, 
                                           num_workers=4)

test_dataset = datasets.CIFAR10(root='./data', 
                                train=False, 
                                download=True, 
                                transform=transform)
test_loader = torch.utils.data.DataLoader(test_dataset, 
                                          batch_size=128, 
                                          shuffle=False, 
                                          num_workers=2)

    
    
def compute_validation_loss(net):
    val_loss = 0.0
    total = 0
    with torch.no_grad():
        for data in test_loader:
            images, labels = data
            images, labels = images.to(device), labels.to(device)
            outputs = net(images)
            loss = nn.CrossEntropyLoss()(outputs, labels)
            val_loss += loss.item()
            total += labels.size(0)
    return val_loss / total


def train(model, optimizer, l1_strength, num_epochs):
    train_losses, val_losses,l1_norms = [],[],[]
    for epoch in range(num_epochs):
        epoch_loss = 0
        for inputs, labels in train_loader:
            inputs, labels = inputs.to(device), labels.to(device)

            # Forward pass
            outputs = model(inputs)
            loss_ce = nn.CrossEntropyLoss()(outputs, labels)

            # L1 penalty
            l1_penalty = 0
            for param in model.parameters():
                l1_penalty += torch.abs(param).sum()

            # Add L1 penalty to the loss
            loss = loss_ce + l1_strength * l1_penalty    

            # Backward and optimize
            optimizer.zero_grad()
            loss.backward()
            optimizer.step()
            epoch_loss += loss.item()
        l1_norms.append(l1_penalty.item())
        train_losses.append(epoch_loss / len(train_loader))
        val_losses.append(compute_validation_loss(model))
        print(f"Epoch: {epoch}, loss: {train_losses[-1]}, val loss: 
        {val_losses[-1]} L1 norm : {l1_norms[-1]}")
    return train_losses, val_losses, l1_norms
    

# Model initialization
vgg16 = models.vgg16(pretrained=True)
initial_state = copy.deepcopy(vgg16.state_dict())
vgg16.to(device)


optimizers = {
    'SGD': optim.SGD(vgg16.parameters(), lr=0.003),
    'RMSProp': optim.RMSprop(vgg16.parameters(), lr=0.003),
    'Adam': optim.Adam(vgg16.parameters(), lr=0.003)
}


def get_num_prunable_params(model, threshold = 1e-5):
    total_params, num_small_params = 0,0
    for name, param in model.named_parameters():
        total_params += param.numel()
        if 'weight' in name or 'bias' in name:
            num_small_params += torch.sum(torch.abs(param) < threshold).item()

    return num_small_params / total_params


loss_curves = {}
l1_norms_curves = {}
validation_loss_curves = {}
num_prunable_params = {}
for name, opt in optimizers.items():
    loss_curves[name], validation_loss_curves[name], l1_norms_curves[name], num_prunable_params[name] = {},{},{},{}
    for lambda1 in [2,0.002]:
        print(f"opt: {name}, lambda: {lambda1}")
        vgg16.load_state_dict(copy.deepcopy(initial_state))  # Reset model
        losses, validation_losses, l1_norms = train(vgg16, opt, lambda1, 25)  
        num_prunable_params[name][lambda1] = get_num_prunable_params(vgg16)
        loss_curves[name][lambda1] = losses
        validation_loss_curves[name][lambda1] = validation_losses
        l1_norms_curves[name][lambda1] = l1_norms
\end{verbatim}

\section{Appendices for Section \ref{sec:eos}}

\subsection{Proof or Proposition \ref{prop:eos}}
\label{app:eos}
  Note that any loss function satisfying \eref{lips} has subgradients bounded above by $L$.  Additionally, as outlined in Section 1 of \citep{boyd2003subgradient}, the deductions derived from the subgradient method hold equivalently for differentiable functions. This equivalence arises from the fact that the only admissible value for the subgradient of a differentiable convex function at any given point is, indeed, its gradient. Accordingly, using the convergence properties of the subgradient methods with constant step sizes from Section 2 of \citep{boyd2003subgradient}, we have that $\lim_ {k\rightarrow\infty} f(x_{k}) - f^{*} \leq \alpha L^{2} < \infty$, hence the result.


\end{document}